\documentclass[11pt]{article}

\usepackage[final]{acl}

\usepackage{times}
\usepackage{latexsym}

\usepackage[T1]{fontenc}

\usepackage[utf8]{inputenc}

\usepackage{microtype}
\usepackage{amsmath}
\usepackage{amssymb}
\usepackage{inconsolata}
\usepackage{booktabs}
\usepackage{graphicx}
\usepackage{colortbl}
\usepackage{xcolor}
\usepackage{caption}
\usepackage{subcaption}
\usepackage{cleveref}
\usepackage{mathtools}
\usepackage{algorithm}
\usepackage{algorithmic}
\usepackage{multirow}
\title{Evolutionary Negative Module Pruning for Better LoRA Merging}

\author{Anda Cao\textsuperscript{1}, Zhuo Gou\textsuperscript{2}, Yi Wang\textsuperscript{1}, Kaixuan Chen\textsuperscript{1,3,4}, Yu Wang\textsuperscript{1} \\ \textbf{Can Wang\textsuperscript{1,3,4}, Mingli Song\textsuperscript{1,3,4}, Jie Song\textsuperscript{2}\thanks{Corresponding author.}} \\
        \textsuperscript{1}College of Computer Science and Technology, Zhejiang University \\
        \textsuperscript{2}School of Software Technology, Zhejiang University \\
        \textsuperscript{3}State Key Laboratory of Blockchain and Security, Zhejiang University \\
        \textsuperscript{4}Hangzhou High-Tech Zone (Binjiang) Institute of Blockchain and Data Security \\
        \texttt{\{caoanda, gouzhuo, y\_w, chenkx, yu.wang\}@zju.edu.cn} \\
        \texttt{\{wcan, brooksong, sjie\}@zju.edu.cn} 
}

\begin{document}
\maketitle
\begin{abstract}
Merging multiple Low-Rank Adaptation (LoRA) experts into a single backbone is a promising approach for efficient multi-task deployment. 
While existing methods strive to alleviate interference via weight interpolation or subspace alignment, they rest upon the implicit assumption that all LoRA matrices contribute constructively to the merged model. 
In this paper, we uncover a critical bottleneck in current merging paradigms: the existence of \textit{negative modules}---specific LoRA layers that inherently degrade global performance upon merging. We propose \textbf{E}volutionary \textbf{N}egative \textbf{M}odule \textbf{P}runing (\textbf{ENMP}), a plug-and-play LoRA pruning method to locate and exclude these detrimental modules prior to merging. By leveraging an evolutionary search strategy, ENMP effectively navigates the discrete, non-differentiable landscape of module selection to identify optimal pruning configurations. 
Extensive evaluations demonstrate that ENMP consistently boosts the performance of existing merging algorithms, achieving a new state-of-the-art across both language and vision domains. Code is available at \url{https://github.com/CaoAnda/ENMP-LoRAMerging}.
\end{abstract}

\section{Introduction}
\label{sec:intro}

Model merging has gained prominence as a scalable paradigm for integrating multiple fine-tuned models into a unified backbone without the prohibitive costs of retraining.
Task Arithmetic (TA) \cite{ilharco2023editing} laid the groundwork for this field by conceptualizing parameter differences as steerable task vectors. Subsequent advancements, such as TIES-Merging \cite{yadav2023ties}, have refined this approach by resolving sign conflicts and pruning redundant parameters. Beyond element-wise aggregation, recent efforts \cite{choi2024revisiting, gargiulo2025task, marczak2025no} have pivoted toward the spectral properties of models, utilizing Singular Value Decomposition (SVD) to harmonize parameter-space conflicts. While these methods make significant advancements for efficient multi-task deployment, they are largely designed for full-parameter fine-tuning.

However, the landscape of model adaptation is shifting alongside the rapid scaling of neural networks \cite{achiam2023gpt, grattafiori2024llama, yang2025qwen3}. As full-rank fine-tuning becomes computationally unsustainable, Parameter-Efficient Fine-Tuning (PEFT) \cite{houlsby2019parameter, li2021prefix, lester2021power, hu2022lora} has emerged as the preferred alternative. Among these, LoRA \cite{hu2022lora} is particularly dominant due to its minimal parameter footprint and robust convergence properties \cite{dettmers2023qlora, zhang2023adaptive}.
With the widespread adoption of LoRA, a single backbone is often required to serve multiple tasks~\cite{wei2022finetuned, sanh2022multitask}, resulting in a rapidly growing number of task-specific adapters. 
Consequently, merging these diverse LoRA experts into a single, cohesive model has become highly attractive for practical deployment. Unfortunately, conventional merging algorithms, which were originally designed for full-parameter adaptations, frequently struggle to account for the unique structural and low-rank constraints inherent to LoRA-adapted models \cite{stoica2025model}.

\begin{figure*}[t]
  \centering
  \begin{minipage}[b]{0.65\textwidth}
    \centering
    \includegraphics[width=\textwidth]{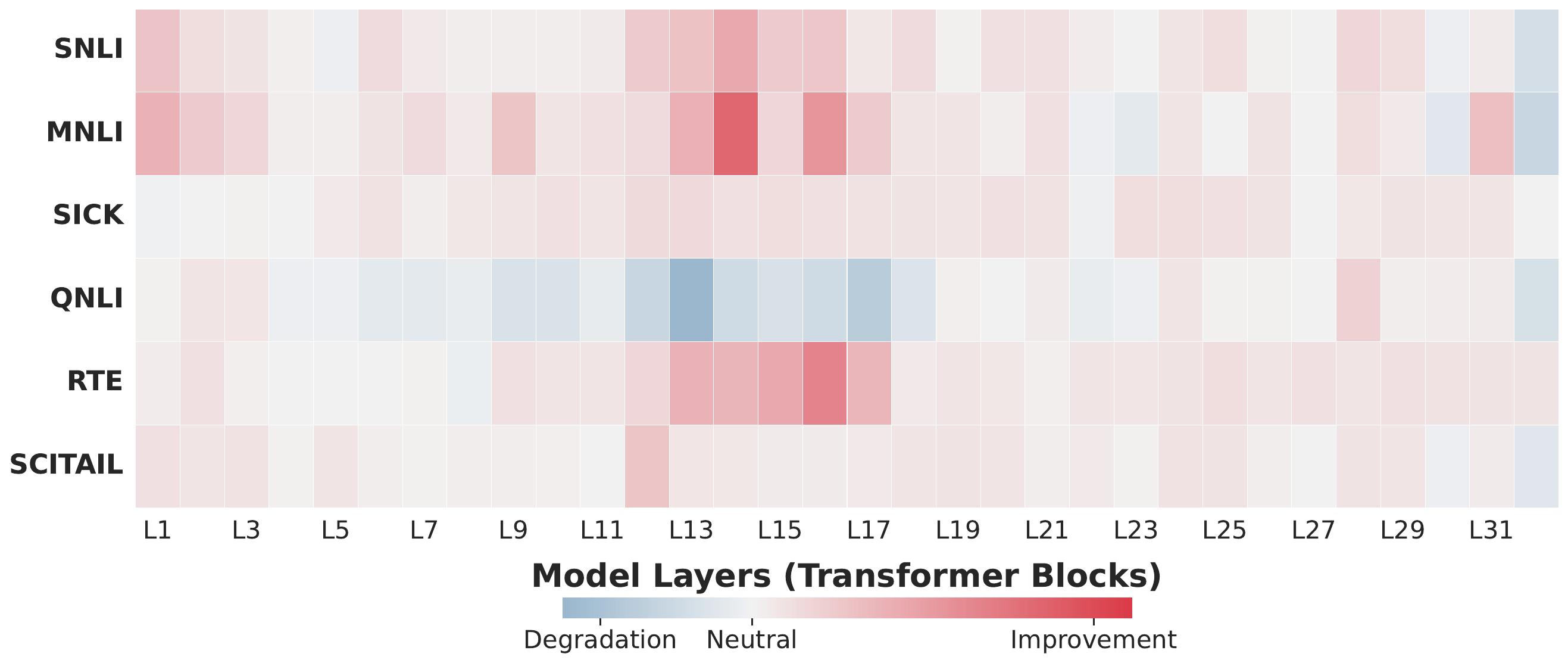}
    \caption{Not all LoRA modules are constructive for merging. 
    In this empirical study, we perform a \textit{leave-one-out} analysis by removing a single LoRA module at a time and merging the rest via Task Arithmetic (TA). 
    The heatmap visualizes the change in average normalized performance of Llama-3-8B across 6 Natural Language Inference (NLI) tasks relative to the baseline (full merge). 
    \textcolor[RGB]{218,59,70}{Red} regions indicate a performance improvement after pruning, while \textcolor[RGB]{154,183,205}{blue} regions indicate a performance drop.}
    \label{fig:heatmap}
  \end{minipage}
  \hfill
  \begin{minipage}[b]{0.32\textwidth}
    \centering
    \includegraphics[width=\textwidth]{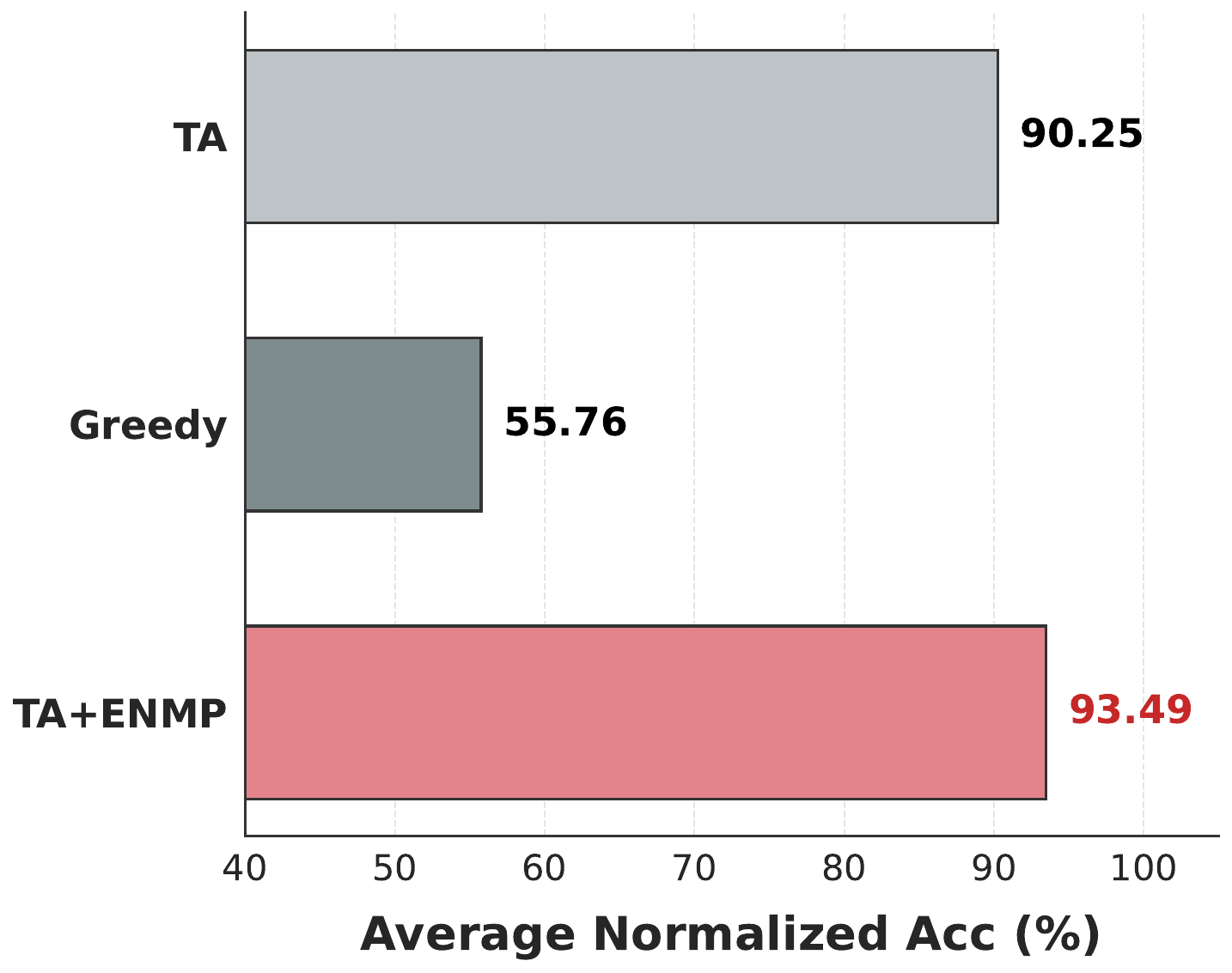}
    \vspace{0pt}
    \caption{Limitations of Greedy Pruning. The Greedy strategy removes all modules that appear detrimental (as suggested in Fig.~\ref{fig:heatmap}). However, this approach ignores cross-layer dependencies, resulting in a degraded accuracy of 55.76\%.}
    \label{fig:greedy_limit}
  \end{minipage}
\end{figure*}

Specifically within the LoRA context, prior works such as KnOTS~\cite{stoica2025model} and CoreSpace~\cite{panariello2025accurate} attribute merging failures to subspace misalignment among independently trained adapters. 
To mitigate this, they project disparate adapters into a shared subspace to enforce compatibility. 
While effective at resolving geometric misalignment, these methods rest upon the idealized assumption that every LoRA module (\textit{i.e.}, the low-rank adaptation matrices) contributes constructively to the merging performance.
In contrast, our empirical study, as shown in~\Cref{fig:heatmap}, uncovers a counter-intuitive phenomenon: removing the entire module in a specific layer from a task-specific LoRA can yield a performance gain in the merged model. 
This suggests that certain modules act as \textit{negative modules}, which exacerbate interference rather than contributing beneficial task-specific information. 
Identifying and pruning these detrimental modules is therefore a critical prerequisite for effective model merging.

However, identifying the optimal subset of modules for pruning presents a formidable challenge, as the impact of a module on merging performance is interdependent: a module characterized as ``negative'' within the full set may become constructive once some detrimental modules are removed, and vice versa. 
This conditional dependency leads to a complex optimization landscape. Consequently, the greedy strategy, which evaluates and prunes modules independently, fails to capture higher-order interactions, resulting in significant performance degradation (\Cref{fig:greedy_limit}).
Furthermore, the search space for the pruning mask suffers from a combinatorial explosion, yielding $2^N$ possible states for $N$ modules, which renders exhaustive search computationally intractable. 
In this work, we propose an evolutionary search approach, termed Evolutionary Negative Module Pruning (ENMP), to efficiently explore the configuration space and locate the optimal pruning mask that maximizes collective performance. 
By enabling the precise pruning of negative modules, ENMP effectively boosts the performance of existing methods, achieving a new state-of-the-art.

To summarize, our contributions are as follows:
\vspace{-0.3em}
\begin{itemize}
    \item We unveil the phenomenon of \textit{negative modules} and demonstrate that pruning them effectively alleviates task interference.\looseness=-1 
    \item We propose {ENMP}, a plug-and-play LoRA pruning method to locate and exclude these detrimental modules prior to merging. 
    \item Extensive evaluations across language and vision domains demonstrate that ENMP consistently boosts the performance of existing merging algorithms.
\end{itemize}
\begin{figure*}[t!]
    \centering
    \includegraphics[width=1.0\linewidth]{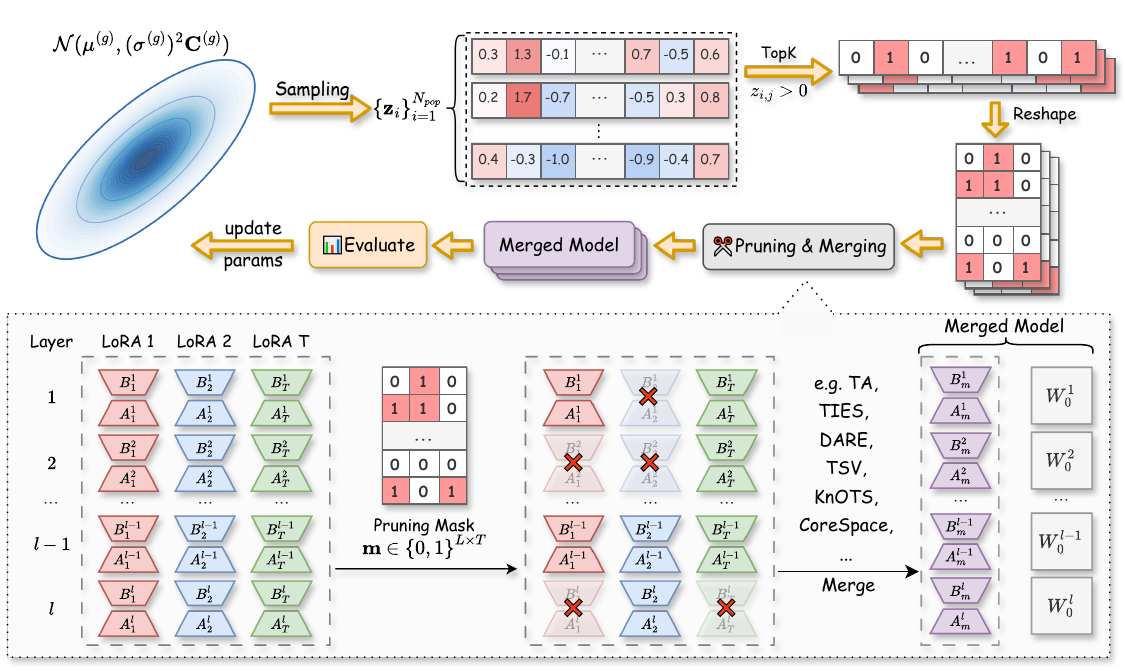}
    \caption{Overview of the proposed framework. The upper part illustrates the optimization loop based on evolutionary strategies (e.g., CMA-ES). Latent variables $\{\mathbf{z}_i\}$ are sampled from a evolving Gaussian distribution $\mathcal{N}(\mu^{(g)}, (\sigma^{(g)})^2\mathbf{C}^{(g)})$ and converted into binary pruning masks via mapping and reshaping operations. The lower part details the \textit{Pruning \& Merging} process: the pruning mask $\mathbf{m}$ is applied to the LoRA adapters to prune negative modules. The remaining adapters are aggregated using existing merging methods (e.g., TIES, DARE) to form the final merged model. The evaluation performance is used to update the distribution parameters iteratively.}
\label{fig:framework}
\end{figure*}
\section{Related Work}

\subsection{Model Merging}
Model merging aims to integrate multiple models independently fine-tuned on different tasks into a single unified model without additional training~\cite{ilharco2023editing}. The field has evolved from simple weight averaging to sophisticated techniques that manipulate task-specific updates. Task Vectors~\cite{ilharco2023editing} established the foundation by demonstrating that weight differences support arithmetic operations for steering model behavior. Building on this, subsequent approaches focus on mitigating interference among conflicting parameters. TIES-Merging~\cite{yadav2023ties} resolves sign conflicts and prunes insignificant updates, while DARE~\cite{yu2024language} and Model Breadcrumbs~\cite{davari2024model} exploit the parameter redundancy in fine-tuned models to enable aggressive sparsification without performance loss. Beyond element-wise manipulation, spectral methods such as TSV~\cite{gargiulo2025task}, Iso-C~\cite{marczak2025no}, and CART~\cite{choi2024revisiting} leverage low-rank properties and singular value alignment to further reduce interference. While effective for full-rank fine-tuned models, these techniques often exhibit instability or suboptimal performance when directly applied to LoRA-adapted models~\cite{tang2024parameterefficient}. 

\subsection{LoRA Merging}

Existing approaches for merging LoRA models can be categorized based on whether they require specialized training strategies. \citet{tang2024parameterefficient} attribute the merging difficulty to increased weight-entanglement and proposes a specialized fine-tuning method involving partial linearization. In contrast, KnOTS~\cite{stoica2025model} presents a gradient-free framework that performs post-hoc alignment of low-rank subspaces using SVD. Building on this, CoreSpace~\cite{panariello2025accurate} further optimizes the paradigm to ensure no information loss. However, these methods implicitly assume that all modules contribute constructively, neglecting the impact of negative modules. To address this, our proposed ENMP departs from the full-retention assumption by introducing an evolutionary search to prune negative modules.

\section{Methodology}
\label{sec:method}

Figure~\ref{fig:framework} illustrates the proposed ENMP framework. Following the preliminaries (Sec.~\ref{subsec:prelim}), we detail the pruning formulation (Sec.~\ref{sec:problem_formulation}) and the optimization process via evolutionary search (Sec.~\ref{sec:cma_es}).

\subsection{Preliminaries}
\label{subsec:prelim}

\noindent
\textbf{Low-Rank Adaptation (LoRA).}
Let $\mathbf{W}_0 \in \mathbb{R}^{d_{\text{out}} \times d_{\text{in}}}$ denote the weight matrix of a dense layer in a pre-trained backbone model. Standard full-rank fine-tuning updates the entire weight matrix, which is computationally expensive and memory-intensive. LoRA~\cite{hu2022lora} hypothesizes that the weight updates have a low intrinsic rank and parameterizes the update $\Delta \mathbf{W}$ by decomposing it into two low-rank matrices $\mathbf{B} \in \mathbb{R}^{d_\text{out} \times r}$ and $\mathbf{A} \in \mathbb{R}^{r \times d_\text{in}}$, where $r \ll \min(d_\text{in}, d_\text{out})$. The forward pass is given by:
\begin{equation}
\mathbf{h} = \mathbf{W}_0 \mathbf{x} + \Delta \mathbf{W} \mathbf{x} = \mathbf{W}_0 \mathbf{x} + \mathbf{B} \mathbf{A} \mathbf{x}.
\end{equation}
In the context of LoRA merging, we consider $T$ diverse tasks. Each task $t \in \{1, \dots, T\}$ is associated with a specific LoRA update $\Delta \mathbf{W}_t$. The collection of these task-specific updates is denoted as $\mathcal{T} = \{ \Delta \mathbf{W}_1, \dots, \Delta \mathbf{W}_T \}$.

\noindent
\textbf{Model Merging.}
Given the set of task-specific experts $\mathcal{T}$ defined above, the goal of model merging is to combine them into a single unified model $\mathbf{W}_{\text{merged}}$, which is capable of handling all tasks simultaneously without retraining. A widely adopted formulation, such as Task Arithmetic~\cite{ilharco2023editing}, computes the weighted sum of task vectors:
\begin{equation}
\mathbf{W}_{\text{merged}} = \mathbf{W}_0 + \lambda \sum_{t=1}^{T} \Delta \mathbf{W}_t,
\label{eq:standard_merge}
\end{equation}
where $\lambda$ is a global scaling coefficient that controls the strength of the merged updates.

\subsection{Merging with Negative Module Pruning}
\label{sec:problem_formulation}

Given the set of LoRA updates $\mathcal{T}$ defined above, standard approaches typically perform a global aggregation across all parameters. We posit that this unselective full aggregation introduces negative modules---specific layers from certain task LoRAs that induce incompatibility and degrade global performance when included. Conversely, removing these modules can recover the model's capabilities.

To implement this, we propose to augment the merging process with a selective pruning mechanism. 
Let $L$ denote the number of Transformer layers in the backbone. 
We define the fundamental pruning unit as the aggregation of LoRA updates for query, key, value, and output projections within a layer, treating them as an indivisible whole to preserve the internal semantic consistency of the attention mechanism. 
Consequently, the total number of pruning units is $N = L \times T$. 
We introduce a binary pruning mask $\mathbf{m} \in \{0, 1\}^{L\times T}$, where $m_{l,t}=1$ indicates that the corresponding LoRA module (\texttt{q\_proj}, \texttt{k\_proj}, \texttt{v\_proj}, \texttt{out\_proj}) at layer $l$ from task $t$ is removed.

We formally define the merging process with module pruning using a generic aggregation function $\Phi(\cdot)$. Let $\mathcal{S}^{(l)}(\mathbf{m}) = \{ \Delta \mathbf{W}_{t}^{(l)} \mid m_{l,t} = 0 \}$ denote the set of retained LoRA modules for layer $l$ based on the pruning mask $\mathbf{m}$ (where 1 indicates pruned). The merged weight is formulated as:
\begin{equation}
  \mathbf{W}_{\text{ENMP}}^{(l)}(\mathbf{m}) = \mathbf{W}_0^{(l)} + \lambda \cdot \Phi \left( \mathcal{S}^{(l)}(\mathbf{m}) \right).
  \label{eq:sta_merge}
\end{equation}
Specifically, we define Task Arithmetic with module pruning as:
\begin{equation}
\Phi_{\text{TA-ENMP}}\left( \mathcal{S}^{(l)}(\mathbf{m}) \right) = \sum_{\mathclap{\Delta \mathbf{W}_t^{(l)} \in \mathcal{S}^{(l)}(\mathbf{m})}} \Delta \mathbf{W}_t^{(l)}.
\label{eq:instantiation}
\end{equation}

Our objective is to find the optimal pruning mask $\mathbf{m}^*$ that maximizes the collective performance across all tasks. Let $\mathcal{D}_\text{val}$ represent the validation data and $\mathcal{M}(\cdot)$ be a comprehensive performance metric (e.g., normalized average accuracy). The optimization problem is formulated as:
\begin{equation}
  \mathbf{m}^* = \operatorname*{arg\,max}_{\mathbf{m} \in \{0, 1\}^{L\times T}} \mathcal{M}\left( \mathbf{W}_{\text{ENMP}}(\mathbf{m}); \mathcal{D}_\text{val} \right).
  \label{eq:objective}
\end{equation}

\subsection{Evolutionary Search with CMA-ES}
\label{sec:cma_es}

Directly searching for the binary pruning mask $\mathbf{m}$ in Eq.~\eqref{eq:objective} is computationally intractable due to the combinatorial explosion of the discrete search space ($2^N$).
More importantly, the decision to prune a specific module is not independent; it heavily relies on the presence or absence of other modules (i.e., cross-layer couplings).
To address these issues, we propose to formulate the pruning task as a search problem within a continuous latent space.

We employ the Covariance Matrix Adaptation Evolution Strategy (CMA-ES)~\cite{hansen2016cma} as our optimizer.
We select CMA-ES specifically for its capability to model the dependencies between decision variables via a covariance matrix, which allows it to capture the complex interactions between LoRA modules that greedy methods overlook.
For clarity and reproducibility, the detailed pseudocode is provided in Appendix~\ref{app:evolutionary_search}.

\vspace{1mm}
\noindent\textbf{Latent Variable \& Mask Mapping.}
Since CMA-ES operates on continuous one-dimensional vectors, we first represent the discrete pruning mask as a flattened vector $\mathbf{m}^{\text{flat}} \in \{0, 1\}^N$.
To bridge the gap between the continuous search space and our discrete objective, we introduce a latent vector $\mathbf{z} \in \mathbb{R}^N$, where each scalar entry $z_j$ serves as a learnable \textit{negativity score} for the $j$-th module.
In this formulation, a higher $z_j$ signifies a stronger tendency for the corresponding module to be pruned.

To translate this continuous score $\mathbf{z}$ into the binary pruning mask, we apply a dynamic thresholding strategy controlled by a maximum pruning ratio $k \in [0, 1)$.
Let $N_{\text{prune}} = \lfloor k \cdot N \rfloor$ denote the budget for allowed removals.
We define $\mathcal{K}$ as the set of indices corresponding to the $N_{\text{prune}}$ largest elements in $\mathbf{z}$.
The pruning mask vector $\mathbf{m}^{\text{flat}}$ is derived by:
\begin{equation}
    m_{j}^{\text{flat}} = 
    \begin{cases}
    1 & \text{if } j \in \mathcal{K} \text{ and } z_j > 0 \\
    0 & \text{otherwise}
    \end{cases}.
    \label{eq:mapping}
\end{equation}

This mapping ensures that the optimization algorithm can explore the continuous landscape of module interference while producing valid discrete masks, which are subsequently reshaped back to the $L \times T$ grid for the physical pruning process described in Section~\ref{sec:problem_formulation}.

\vspace{1mm}
\noindent\textbf{Conservative Initialization.} 
We initialize the mean vector of the CMA-ES population to a uniform value of $-1$ (i.e., $\boldsymbol{\mu}^{(0)} = -\mathbf{1}$). 
Since our mapping requires $z_j > 0$ to trigger pruning, this negative initialization ensures that the search begins from the \textit{fully merging} state (where all $m_{l, t}=0$).

\vspace{1mm}
\noindent\textbf{Search Process.}
The optimization proceeds iteratively over $G$ generations. In each generation $g$, the algorithm executes three key steps:

\begin{itemize}
  \item \textbf{Sampling:} We generate a population of $N_\text{pop}$ candidate latent vectors $\{\mathbf{z}_1, \dots, \mathbf{z}_{N_\text{pop}}\}$ from a multivariate normal distribution:
\begin{equation}
  \mathbf{z}_i \sim \mathcal{N}(\boldsymbol{\mu}^{(g)}, (\sigma^{(g)})^2 \mathbf{C}^{(g)}), i = 1, \dots, N_\text{pop},
\end{equation}
where $\boldsymbol{\mu}^{(g)}$ represents the current estimate of the optimal negativity scores, $\sigma^{(g)}$ is the step size, and $\mathbf{C}^{(g)} \in \mathbb{R}^{N \times N}$ is the covariance matrix determining the geometric shape and variable dependencies of the search distribution.\looseness=-1

\item \textbf{Evaluation:} For each candidate $\mathbf{z}_i$, we generate the binary mask $\mathbf{m}_i$ via Eq.~\eqref{eq:mapping}. We then construct the merged model $\mathbf{W}_{\text{ENMP}}(\mathbf{m}_i)$ and evaluate its fitness (e.g., validation normalized accuracy) to obtain a score $\mathcal{M}_i$.

\item \textbf{Update:} The candidates are sorted by their fitness scores. 
The mean $\boldsymbol{\mu}^{(g+1)}$ is updated via a weighted average of the top-performing candidates. 
Simultaneously, the step-size $\sigma^{(g+1)}$ and covariance matrix $\mathbf{C}^{(g+1)}$ are adapted to control the exploration magnitude and capture the dependencies between modules.
\end{itemize}

This ability to model variable correlations allows ENMP to navigate the complex, non-separable landscape of module interference.

\vspace{1mm}
\noindent\textbf{Final Model Construction \& Complexity.}
We track the candidate latent vector $\mathbf{z}_{\text{best}}$ that achieves the highest fitness score throughout the evolutionary process. 
Upon convergence or exhaustion of the generation budget, the optimal binary mask is derived as $\mathbf{m}^* = \text{Mapping}(\mathbf{z}_{\text{best}})$. 
The final unified model is constructed by applying this mask to the merging formulation in Eq.~\eqref{eq:sta_merge}:
\begin{equation}
  \mathbf{W}_{\text{final}} = \mathbf{W}_{\text{ENMP}}(\mathbf{m}^*) = \mathbf{W}_0 + \lambda \cdot \Phi \left( \mathcal{S}(\mathbf{m}^*) \right).
  \label{eq:final_construction}
\end{equation}

Notably, this optimization process represents a \textit{one-time offline cost}. 
The merged model retains the identical architectural footprint as the original pre-trained model, incurring zero additional inference overhead in terms of latency or memory usage.

\section{Experiments}
\label{sec:experiments}

\begin{table*}[t!]
\renewcommand{\arraystretch}{0.92}
\footnotesize
\centering
\resizebox{\textwidth}{!}{%
\begin{tabular}{llllllllc}
\toprule
\textbf{Method} & \textbf{SNLI} & \textbf{MNLI} & \textbf{SICK} & \textbf{QNLI} & \textbf{RTE} & \textbf{SciTail} & \textbf{Avg.} & \textbf{$\Delta$} \\
\midrule
TA & 93.60 & \textbf{95.29} & 88.00 & 68.73 & 99.19 & 96.69 & 90.25 & - \\
\rowcolor{gray!15}\textbf{TA + ENMP} & $\textbf{95.93}_{\pm 0.28}$ & $94.32_{\pm 0.43}$ & $\textbf{91.59}_{\pm 1.78}$ & $\textbf{80.40}_{\pm 0.60}$ & $\textbf{101.34}_{\pm 0.46}$ & $\textbf{97.37}_{\pm 0.13}$ & $\textbf{93.49}_{\pm 0.26}$ & \textbf{+3.24} \\
\midrule
TIES & 94.86 & 96.71 & 80.79 & 71.54 & \textbf{100.00} & 96.00 & 89.99 & - \\
\rowcolor{gray!15}\textbf{TIES + ENMP} & $\textbf{96.57}_{\pm 0.27}$ & $\textbf{98.23}_{\pm 0.63}$ & $\textbf{92.73}_{\pm 0.81}$ & $\textbf{94.95}_{\pm 0.67}$ & $99.46_{\pm 0.47}$ & $\textbf{96.39}_{\pm 0.42}$ & $\textbf{96.39}_{\pm 0.45}$ & \textbf{+6.40} \\
\midrule
DARE & 94.50 & 96.87 & 77.68 & 72.44 & 97.58 & 96.15 & 89.20 & - \\
\rowcolor{gray!15}\textbf{DARE + ENMP} & $\textbf{96.09}_{\pm 0.58}$ & $\textbf{97.77}_{\pm 0.71}$ & $\textbf{92.34}_{\pm 0.66}$ & $\textbf{94.94}_{\pm 0.74}$ & $\textbf{98.66}_{\pm 1.23}$ & $\textbf{97.22}_{\pm 0.26}$ & $\textbf{96.17}_{\pm 0.14}$ & \textbf{+6.97} \\
\midrule
TSV & 95.37 & \textbf{95.13} & 88.85 & 76.82 & \textbf{101.61} & 97.56 & 92.56 & - \\
\rowcolor{gray!15}\textbf{TSV + ENMP} & $\textbf{96.45}_{\pm 0.75}$ & $94.49_{\pm 0.70}$ & $\textbf{94.40}_{\pm 2.21}$ & $\textbf{92.19}_{\pm 0.87}$ & $100.00_{\pm 1.40}$ & $\textbf{97.58}_{\pm 0.28}$ & $\textbf{95.85}_{\pm 0.37}$ & \textbf{+3.29} \\
\midrule
KnOTS & 89.29 & 94.08 & 89.67 & 83.63 & 100.81 & \textbf{97.37} & 92.47 & - \\
\rowcolor{gray!15}\textbf{KnOTS + ENMP} & $\textbf{95.11}_{\pm 0.37}$ & $\textbf{97.48}_{\pm 1.22}$ & $\textbf{96.60}_{\pm 0.89}$ & $\textbf{95.97}_{\pm 1.36}$ & $\textbf{101.34}_{\pm 1.23}$ & $97.24_{\pm 0.44}$ & $\textbf{97.29}_{\pm 0.29}$ & \textbf{+4.82} \\
\midrule
CoreSpace & 95.84 & \textbf{95.74} & 89.25 & 83.97 & \textbf{102.42} & 97.86 & 94.18 & - \\
\rowcolor{gray!15}\textbf{CoreSpace + ENMP} & $\textbf{97.56}_{\pm 0.02}$ & $93.37_{\pm 0.79}$ & $\textbf{96.00}_{\pm 0.71}$ & $\textbf{93.69}_{\pm 1.25}$ & $101.34_{\pm 0.46}$ & $\textbf{98.39}_{\pm 0.46}$ & $\textbf{96.73}_{\pm 0.13}$ & \textbf{+2.55} \\
\bottomrule
\end{tabular}%
}
\caption{Performance comparison on NLP benchmark. We report normalized accuracy (\%) as mean $\pm$ std over 3 runs. See Appendix~\ref{app:nlp_results} for absolute accuracy. $\Delta$ denotes the improvement in average accuracy. \textbf{Bold} marks the best result per group. ENMP consistently outperforms all baselines, including advanced alignment methods.}
\label{tab:main_results_nlp}
\end{table*}

\begin{table*}[t!]
\centering
\resizebox{\textwidth}{!}{%
\begin{tabular}{llllllllllc}
\toprule
\textbf{Method} & \textbf{Cars} & \textbf{DTD} & \textbf{EuroSAT} & \textbf{GTSRB} & \textbf{MNIST} & \textbf{RESISC} & \textbf{SUN397} & \textbf{SVHN} & \textbf{Avg.} & \textbf{$\Delta$} \\
\midrule
TA & \textbf{81.99} & 73.72 & 49.31 & \textbf{42.17} & 53.01 & 71.48 & 95.39 & 41.22 & 63.54 & - \\
\rowcolor{gray!15}\textbf{TA + ENMP} & $81.60_{\pm 0.65}$ & $\textbf{75.18}_{\pm 0.75}$ & $\textbf{55.02}_{\pm 0.47}$ & $41.19_{\pm 1.49}$ & $\textbf{58.08}_{\pm 1.96}$ & $\textbf{71.85}_{\pm 0.23}$ & $\textbf{96.09}_{\pm 0.20}$ & $\textbf{42.63}_{\pm 0.96}$ & $\textbf{65.21}_{\pm 0.34}$ & \textbf{+1.67} \\
\midrule
TIES & \textbf{82.60} & 73.17 & 50.77 & 36.68 & 57.48 & 69.67 & 95.35 & 42.97 & 63.59 & - \\
\rowcolor{gray!15}\textbf{TIES + ENMP} & $81.53_{\pm 0.40}$ & $\textbf{74.24}_{\pm 0.92}$ & $\textbf{58.16}_{\pm 2.22}$ & $\textbf{37.93}_{\pm 0.53}$ & $\textbf{65.58}_{\pm 0.73}$ & $\textbf{70.48}_{\pm 0.22}$ & $\textbf{95.93}_{\pm 0.53}$ & $\textbf{46.82}_{\pm 2.40}$ & $\textbf{66.33}_{\pm 0.34}$ & \textbf{+2.74} \\
\midrule
DARE & \textbf{82.43} & 72.99 & 49.91 & 37.65 & 56.71 & 69.74 & 95.10 & 44.50 & 63.63 & - \\
\rowcolor{gray!15}\textbf{DARE + ENMP} & $82.04_{\pm 0.22}$ & $\textbf{75.54}_{\pm 0.55}$ & $\textbf{58.89}_{\pm 2.01}$ & $\textbf{39.29}_{\pm 0.42}$ & $\textbf{66.90}_{\pm 0.32}$ & $\textbf{70.92}_{\pm 0.47}$ & $\textbf{96.11}_{\pm 0.12}$ & $\textbf{47.58}_{\pm 1.50}$ & $\textbf{67.16}_{\pm 0.04}$ & \textbf{+3.53} \\
\midrule
TSV & \textbf{83.59} & 75.45 & 52.60 & 44.85 & 59.53 & 73.39 & 95.25 & 49.29 & 66.74 & - \\
\rowcolor{gray!15}\textbf{TSV + ENMP} & $82.75_{\pm 0.46}$ & $\textbf{75.48}_{\pm 1.83}$ & $\textbf{62.56}_{\pm 3.81}$ & $\textbf{46.46}_{\pm 0.86}$ & $\textbf{69.49}_{\pm 1.76}$ & $\textbf{73.90}_{\pm 0.90}$ & $\textbf{96.22}_{\pm 0.42}$ & $\textbf{53.12}_{\pm 0.55}$ & $\textbf{70.00}_{\pm 0.21}$ & \textbf{+3.26} \\
\midrule
KnOTS & 82.74 & 72.63 & 47.06 & 44.14 & 61.59 & 71.25 & 93.58 & 49.22 & 65.28 & - \\
\rowcolor{gray!15}\textbf{KnOTS + ENMP} & $\textbf{82.98}_{\pm 0.35}$ & $\textbf{75.64}_{\pm 0.16}$ & $\textbf{55.06}_{\pm 0.72}$ & $\textbf{44.76}_{\pm 1.79}$ & $\textbf{80.46}_{\pm 1.53}$ & $\textbf{71.38}_{\pm 1.62}$ & $\textbf{95.41}_{\pm 0.74}$ & $\textbf{60.91}_{\pm 2.81}$ & $\textbf{70.82}_{\pm 0.31}$ & \textbf{+5.54} \\
\midrule
CoreSpace & 82.89 & \textbf{85.03} & 53.16 & \textbf{84.30} & 71.00 & 84.34 & 97.51 & 53.55 & 76.47 & - \\
\rowcolor{gray!15}\textbf{CoreSpace + ENMP} & $\textbf{84.14}_{\pm 0.12}$ & $82.54_{\pm 2.79}$ & $\textbf{67.15}_{\pm 1.58}$ & $81.40_{\pm 1.55}$ & $\textbf{76.28}_{\pm 1.26}$ & $\textbf{84.98}_{\pm 0.64}$ & $\textbf{97.80}_{\pm 0.46}$ & $\textbf{56.06}_{\pm 2.11}$ & $\textbf{78.79}_{\pm 0.22}$ & \textbf{+2.32} \\
\bottomrule
\end{tabular}%
}
\caption{Results on Vision benchmarks (ViT-B/32). We report normalized accuracy (\%) as mean $\pm$ std over 3 runs. See Appendix~\ref{app:cv_results} for absolute accuracy. $\Delta$ denotes the improvement in average accuracy. \textbf{Bold} marks the best result per group. ENMP demonstrates robust improvements across diverse visual recognition tasks.}
\label{tab:main_results_cv}
\end{table*}

\subsection{Experimental Setup}
\paragraph{Benchmarks and Metrics.}
To ensure fair comparison and reproducibility, we follow the experimental protocols established in prior studies~\cite{stoica2025model,panariello2025accurate}, conducting evaluations across two domains: Natural Language Processing (NLP) and Computer Vision (CV).

For the NLP benchmark, we focus on 6 Natural Language Inference (NLI) tasks: \texttt{SNLI}~\cite{bowman2015large}, \texttt{MNLI}~\cite{williams2018broad}, \texttt{SICK}~\cite{marelli-etal-2014-sick}, \texttt{QNLI}~\cite{wang-etal-2018-glue}, \texttt{RTE}~\cite{wang-etal-2018-glue}, and \texttt{SciTail}~\cite{khot2018scitail}.
For the CV benchmark, we use 8 image classification datasets: \texttt{Cars}~\cite{krause20133d}, \texttt{DTD}~\cite{cimpoi2014describing}, \texttt{EuroSAT}~\cite{helber2019eurosat}, \texttt{GTSRB}~\cite{stallkamp2011german}, \texttt{MNIST}~\cite{lecun-mnisthandwrittendigit-2010}, \texttt{RESISC45}~\cite{cheng2017remote}, \texttt{SUN397}~\cite{xiao2016sun}, and \texttt{SVHN}~\cite{netzer2011reading}.\looseness=-1

To account for the varying difficulty levels across diverse tasks, we adopt \textit{Normalized Accuracy} as our primary evaluation metric. 
This metric calibrates the merged model's performance relative to the single-task experts, ensuring a balanced comparison. 
For completeness, the detailed absolute accuracy scores for all individual tasks are reported in Appendix~\ref{app:additional_results}. 
Formally, for a given task $t$, let $\text{Acc}_{\text{merged}}^{(t)}$ denote the accuracy of the merged model and $\text{Acc}_{\text{expert}}^{(t)}$ be the accuracy of the single-task expert. 
The normalized accuracy is defined as:
\begin{equation}
    \text{NormAcc}^{(t)} = \frac{\text{Acc}_{\text{merged}}^{(t)}}{\text{Acc}_{\text{expert}}^{(t)}}.
\end{equation}
This metric reflects how well the merged model retains task-specific capabilities relative to the dedicated experts. 
We report both the individual scores and the benchmark average to highlight task-specific variances and global trends.

\paragraph{Models and Architectures.}
We conduct experiments using two distinct architectures to demonstrate the generality of our approach across modalities:
\textit{(1) Large Language Models}: We employ \texttt{Llama-3-8B}~\cite{grattafiori2024llama}, a representative decoder-only model, to evaluate performance on NLI tasks.
\textit{(2) Vision Encoders}: For cross-modal validation, we use the \texttt{ViT-B/32} variant of the CLIP vision encoder~\cite{radford2021learning}. 

\paragraph{Baselines.}
We validate the efficacy of ENMP by integrating it with representative state-of-the-art parameter merging methods, including Task Arithmetic (TA)~\cite{ilharco2023editing}, TIES-Merging~\cite{yadav2023ties}, DARE~\cite{yu2024language}, TSV~\cite{gargiulo2025task}, KnOTS~\cite{stoica2025model}, and CoreSpace~\cite{panariello2025accurate}. Detailed formulations of these baselines are provided in Appendix~\ref{app:baseline_formulations}. 
By treating these methods as foundational baselines, we demonstrate that ENMP serves as a versatile plug-and-play module that consistently enhances their performance by effectively locating and removing negative modules.

\paragraph{Implementation Details.}
To ensure a rigorous comparison, we use the pre-trained LoRA checkpoints provided by \citet{stoica2025model}. 
Consistent with existing studies, we apply LoRA adapters to all weight matrices in the attention mechanism (\texttt{q\_proj}, \texttt{k\_proj}, \texttt{v\_proj}, \texttt{out\_proj}), setting the rank $r=16$ and scaling factor $\alpha=16$.
For the optimization phase, we use the same held-out validation sets as \citet{panariello2025accurate}.
The pruning mask $\mathbf{m}$ is optimized via CMA-ES with a population size $N_\text{pop}=16$ for 60 generations. 
Unless otherwise stated, we configure the search with an initial step size $\sigma=0.5$ and a maximum pruning ratio $k=0.2$. 
Detailed hyperparameter configurations are provided in Appendix~\ref{app:hyperparams}.
All experiments are conducted on 8 NVIDIA RTX 4090 GPUs, where the candidate evaluations per CMA-ES generation are distributed across GPUs in parallel.

\subsection{Main Results}
\label{subsec:main_results}

We evaluate the proposed framework on both NLP and Vision benchmarks to verify its effectiveness and generalizability.

\paragraph{Performance on NLP Benchmark.}
Table~\ref{tab:main_results_nlp} presents the comprehensive evaluation results on the NLP benchmark.
The empirical evidence strongly supports our hypothesis that selectively removing negative modules enhances the performance of merged language models.
Detailed observations are discussed below.

\textit{(1) Universal enhancement.}
ENMP consistently improves performance across all baselines, regardless of the underlying merging strategy.
It provides notable gains for simple methods (e.g., +3.24\% for TA) while propelling advanced baselines like KnOTS to a new state-of-the-art accuracy of 97.29\%.
This universality indicates that negative modules are a pervasive bottleneck in LoRA merging, spanning from element-wise aggregation to spectral alignment methods.\looseness=-1

\textit{(2) Synergy with sparsification methods.}
We observe substantial improvements when integrating ENMP with parameter-level sparsification methods.
Specifically, it boosts TIES by +6.40\% and DARE by +6.97\%.
These gains confirm that while fine-grained pruning reduces intra-module redundancy, it neglects inter-module interference.
ENMP effectively alleviates interference at the modular level.

\begin{figure}[t!]
    \centering
    \includegraphics[width=1\linewidth]{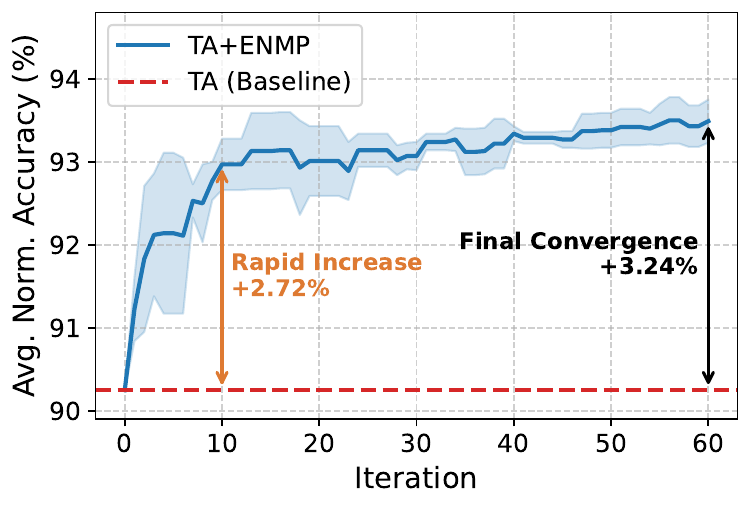}
    \caption{Optimization trajectory of ENMP. We track the test set average normalized accuracy of the pruning mask selected based on validation performance at each generation. The shaded region represents the standard deviation across three independent runs. }
    \label{fig:enmp_convergence_curve}
\end{figure}

\textit{(3) Critical recovery on sensitive tasks.}
Notably, the performance recovery on QNLI is particularly striking.
While baseline methods like TIES and DARE suffer from severe performance degradation on this dataset, ENMP achieves a remarkable gain of over +20\%, restoring the normalized accuracy to approximately 95\%.
This drastic improvement suggests that task interference is not uniformly distributed but can be catastrophic for specific sensitive tasks.
It demonstrates that structural conflicts can be effectively resolved by physically removing the specific modules that induce negative transfer.

\paragraph{Cross-Modal Generalizability.}
To further verify that our method addresses the fundamental problem of module interference rather than overfitting to specific language architectures, we extend our evaluation to the vision domain (Table~\ref{tab:main_results_cv}).
Consistent with the NLP findings, ENMP yields robust improvements across diverse image recognition tasks~(e.g., +5.54\% for KnOTS).
These results indicate that the phenomenon of negative modules is pervasive and our framework is modality-agnostic.

\subsection{Search Efficiency}
The search efficiency of the proposed framework directly determines its practical viability.
Using TA as the representative baseline, we investigate the convergence behavior of ENMP to evaluate how quickly it finds effective pruning masks.
In Figure~\ref{fig:enmp_convergence_curve}, we plot the test set performance of the candidate mask that achieves the highest accuracy on the validation set at each generation.
As illustrated, the optimization trajectory exhibits a steep accuracy ascent during the early phase---the method achieves a rapid performance boost of +2.72\% within the first 10 iterations (approximately 23 minutes on our experimental setup), indicating that the evolutionary algorithm efficiently identifies the most significant negative modules.
The complete 60-generation search converges to a final gain of +3.24\% in approximately 2.3 hours, suggesting that the majority of the gain is captured early in the search.
Notably, the test set accuracy improves in tandem with the optimization on the validation set, indicating that ENMP navigates the combinatorial search space without overfitting to the validation data.

\subsection{Synergy with Subspace Alignment}
\label{sub_align}

\begin{figure}[t]
    \centering
\includegraphics[width=0.981\linewidth]{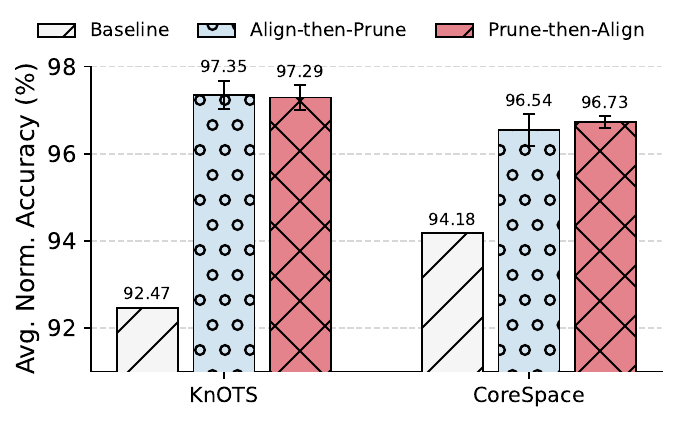}
    \caption{Synergy between ENMP and Subspace Alignment. 
    \textit{Prune-then-Align} consistently outperforms or matches the reverse order by preventing negative modules from polluting the shared subspace.}
    \label{fig:order_comparison}
\end{figure}

We investigate the interaction between ENMP and subspace alignment methods (KnOTS~\cite{stoica2025model}, CoreSpace~\cite{panariello2025accurate}) by comparing two execution orders: \textit{Align-then-Prune} and \textit{Prune-then-Align}. In Figure~\ref{fig:order_comparison}, \textit{Prune-then-Align} generally yields superior performance, boosting CoreSpace by +0.19\%.
We attribute this to the sensitivity of subspace construction. In \textit{Align-then-Prune}, negative modules participate in the basis construction via SVD, ``polluting'' the shared subspace with interference directions. Conversely, \textit{Prune-then-Align} derives the subspace solely from constructive modules, ensuring a cleaner reference basis for alignment. While KnOTS shows robustness to ordering (around 97.3\%), early pruning remains theoretically preferable for reducing the computational overhead of the SVD step.

\begin{table}[t!]
\centering
\resizebox{1.0\linewidth}{!}{
\begin{tabular}{l c c}
\toprule
\textbf{Method} & \textbf{Avg. Norm. Acc. (\%)} & \textbf{$\Delta$} \\
\midrule
Task Arithmetic (TA) & 90.25 & - \\
TA + Random Pruning & 89.10 \small{$\pm$ 1.19} & \textcolor{red}{-1.15} \\
\textbf{TA + ENMP (Ours)} & \textbf{93.49} \small{$\pm$ 0.26} & \textcolor{green!60!black}{\textbf{+3.24}} \\
\bottomrule
\end{tabular}
}
\caption{Comparison with Random Pruning on the NLP benchmark. $\Delta$ denotes the change in average normalized accuracy relative to the Task Arithmetic baseline.}
\label{tab:random_ablation}
\end{table}

\section{Ablation Study}
\label{sec:ablation}

We validate the necessity of evolutionary search and the robustness to pruning constraints, employing Task Arithmetic (TA) as the representative baseline to analyze the impact of ENMP.

\subsection{Necessity of Evolutionary Search}
To investigate whether performance gains stem from precise pruning or mere sparsity, we compare ENMP with a \textit{Random Pruning} baseline (matching approximately 16.7\% sparsity resulting from ENMP's search) on the NLP benchmark. As shown in Table~\ref{tab:random_ablation}, random pruning degrades accuracy to 89.10\% representing a 1.15\% drop from the TA baseline, with high variance ($\pm$1.19\%), indicating that interference is non-uniformly distributed and blind removal risks discarding essential knowledge. In contrast, ENMP achieves 93.49\% with negligible variance ($\pm$0.26\%), confirming that the advantage derives from the precise localization of interfering components rather than sparsity alone.

\subsection{Robustness to Pruning Constraints}
\label{ablation:topk}

\begin{figure}[t]
    \centering
    \includegraphics[width=1\linewidth]{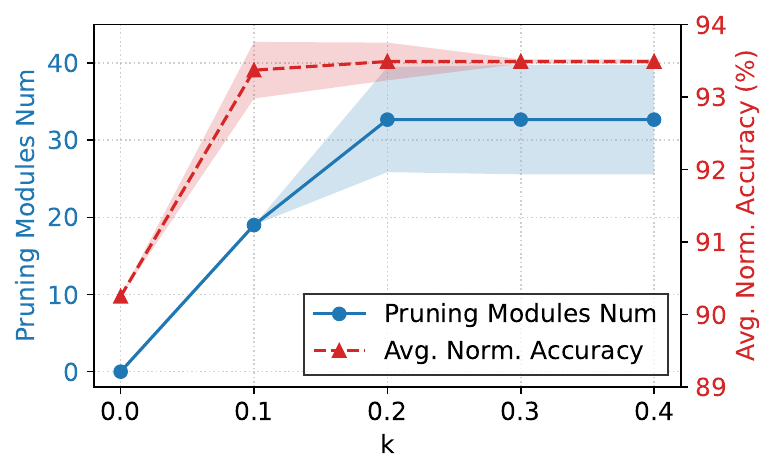}
    \caption{Sensitivity to Maximum Pruning Ratio ($k$). 
Results (mean $\pm$ std over three runs) show increasing $k$ leads to saturated module removal (\textcolor[RGB]{31, 119, 180}{left}) and stable accuracy (\textcolor[RGB]{214, 39, 40}{right}), confirming ENMP's \textit{adaptive sparsity}.}
    \label{fig:topk_curve}
\end{figure}

We empirically analyze ENMP's sensitivity to the maximum pruning ratio $k$ ($0.0$ to $0.4$) in Figure~\ref{fig:topk_curve}.
Even a modest relaxation from $k{=}0.0$ to $k{=}0.1$ triggers a sharp accuracy boost (from $90.25\%$ to $93.37\%$), indicating that removing even a few negative modules yields substantial benefits.
For looser constraints ($k \ge 0.2$), ENMP exhibits \textit{adaptive sparsity}: the number of removed modules plateaus at approximately 32 rather than blindly filling the allowed budget.
This stability suggests that $k$ serves merely as a flexible upper bound; providing a sufficient margin allows the algorithm to autonomously converge to the optimal sparsity level without fine-grained tuning.

\begin{table}[t!]
\centering
\resizebox{1.0\linewidth}{!}{
    \begin{tabular}{ccccc}
    \toprule
    \textbf{Samples / Task} & \textbf{64} & \textbf{128} & \textbf{256} & \textbf{Full Val} \\
    \midrule
    TA + ENMP (Ours) & 91.17 & 91.86 & 92.24 & 93.49 \\
    $\Delta$ (vs. TA) & +0.92 & +1.61 & +1.99 & +3.24 \\
    \bottomrule
    \end{tabular}
}
\caption{Sensitivity to Validation Set Size on NLP Benchmark. Average normalized accuracy (\%) with varying numbers of samples per task. $\Delta$ denotes improvement over the TA baseline (90.25\%).}
\label{tab:data_efficiency}
\end{table}

\subsection{Sensitivity to Validation Set Size}
\label{ablation:datasize}

To evaluate how much validation data ENMP requires for reliable fitness evaluation, we vary the number of samples per task from 64 to the full validation set on the NLP benchmark. As shown in Table~\ref{tab:data_efficiency}, ENMP exhibits strong data efficiency. With as few as 64 samples per task, it already achieves 91.17\% accuracy, surpassing the TA baseline (90.25\%) by +0.92\%. Performance further improves consistently as more data becomes available.
This result suggests that the interference signal captured by ENMP is structural rather than statistical. Even under high sampling variance, ENMP can consistently identify negative modules, reducing the reliance on large-scale external validation sets.

\section{Conclusion}
\label{sec:conclusion}

In this work, we challenge the idealized assumption in LoRA merging that all task-specific parameters contribute constructively to the merged model. We uncover the existence of \textit{negative modules}, specific LoRA layers that introduce interference and degrade multi-task performance. To address this, we propose ENMP, a novel framework that utilizes the evolutionary search algorithm to solve negative module pruning as a combinatorial optimization problem. Extensive experiments across NLP and vision benchmarks demonstrate that ENMP serves as a versatile, plug-and-play enhancement, consistently boosting the performance of existing state-of-the-art merging algorithms (including TIES, KnOTS, and CoreSpace). Our findings highlight that in the era of massive model composability, subtractive mechanisms are just as critical as additive ones for achieving optimal model merging. 

\section*{Limitations}
\label{sec:limitations}

While ENMP effectively enhances model merging performance, we identify two primary limitations to be addressed in future work.

\paragraph{Offline Computational Overhead.}
Unlike instant merging methods such as Task Arithmetic, ENMP involves an iterative evolutionary search. Although this incurs a one-time computational cost during the merging phase, it is important to note that the merged model retains the exact architecture of the backbone, incurring \textit{zero additional overhead during inference}.
Our experiments show efficient convergence within 60 generations (Figure~\ref{fig:enmp_convergence_curve}). However, applying this search to extremely large-scale settings (e.g., 70B models with hundreds of tasks) remains a challenge that may necessitate more sample-efficient optimization strategies.

\paragraph{Requirement for Validation Data.}
ENMP utilizes a validation set ($\mathcal{D}_\text{val}$) to compute fitness scores for the evolutionary algorithm. While this dependency allows for precise localization of negative modules, it assumes the availability of representative labeled data. In scenarios strictly requiring data-free merging, this prerequisite constitutes a constraint. Nevertheless, given that many state-of-the-art baselines also benefit from validation data for hyperparameter tuning (e.g., scaling factors), we consider this a justifiable trade-off for the significant performance gains observed.

\section*{Acknowledgments}
This work was supported by the National Natural Science Foundation of China (62576305, 62506330), the Ningbo Project of Leading Youth Talents for S\&T Innovation (2025QL052), and the Zhejiang Provincial Natural Science Foundation of China (LQN26F020007).

\bibliography{custom}

@inproceedings{
panariello2025accurate,
title={Accurate and Efficient Low-Rank Model Merging in Core Space},
author={Aniello Panariello and Daniel Marczak and Simone Magistri and Angelo Porrello and Bart{\l}omiej Twardowski and Andrew D. Bagdanov and Simone Calderara and Joost van de Weijer},
booktitle={The Thirty-ninth Annual Conference on Neural Information Processing Systems},
year={2025},
url={https://openreview.net/forum?id=y1z7SAS8q8}
}

@article{yadav2023ties,
  title={Ties-merging: Resolving interference when merging models},
  author={Yadav, Prateek and Tam, Derek and Choshen, Leshem and Raffel, Colin A and Bansal, Mohit},
  journal={Advances in Neural Information Processing Systems},
  volume={36},
  pages={7093--7115},
  year={2023}
}

@inproceedings{davari2024model,
  title={Model breadcrumbs: Scaling multi-task model merging with sparse masks},
  author={Davari, MohammadReza and Belilovsky, Eugene},
  booktitle={European Conference on Computer Vision},
  pages={270--287},
  year={2024},
  organization={Springer}
}

@inproceedings{
ilharco2023editing,
title={Editing models with task arithmetic},
author={Gabriel Ilharco and Marco Tulio Ribeiro and Mitchell Wortsman and Ludwig Schmidt and Hannaneh Hajishirzi and Ali Farhadi},
booktitle={The Eleventh International Conference on Learning Representations },
year={2023},
url={https://openreview.net/forum?id=6t0Kwf8-jrj}
}

@inproceedings{gargiulo2025task,
  title={Task singular vectors: Reducing task interference in model merging},
  author={Gargiulo, Antonio Andrea and Crisostomi, Donato and Bucarelli, Maria Sofia and Scardapane, Simone and Silvestri, Fabrizio and Rodola, Emanuele},
  booktitle={Proceedings of the Computer Vision and Pattern Recognition Conference},
  pages={18695--18705},
  year={2025}
}

@inproceedings{
marczak2025no,
title={No Task Left Behind: Isotropic Model Merging with Common and Task-Specific Subspaces},
author={Daniel Marczak and Simone Magistri and Sebastian Cygert and Bart{\l}omiej Twardowski and Andrew D. Bagdanov and Joost van de Weijer},
booktitle={Forty-second International Conference on Machine Learning},
year={2025},
url={https://openreview.net/forum?id=RBZpAa27ls}
}

@inproceedings{
tang2024parameterefficient,
title={Parameter-Efficient Multi-Task Model Fusion with Partial Linearization},
author={Anke Tang and Li Shen and Yong Luo and Yibing Zhan and Han Hu and Bo Du and Yixin Chen and Dacheng Tao},
booktitle={The Twelfth International Conference on Learning Representations},
year={2024},
url={https://openreview.net/forum?id=iynRvVVAmH}
}

@inproceedings{
stoica2025model,
title={Model merging with {SVD} to tie the Knots},
author={George Stoica and Pratik Ramesh and Boglarka Ecsedi and Leshem Choshen and Judy Hoffman},
booktitle={The Thirteenth International Conference on Learning Representations},
year={2025},
url={https://openreview.net/forum?id=67X93aZHII}
}

@inproceedings{
    hu2022lora,
    title={Lo{RA}: Low-Rank Adaptation of Large Language Models},
    author={Edward J Hu and yelong shen and Phillip Wallis and Zeyuan Allen-Zhu and Yuanzhi Li and Shean Wang and Lu Wang and Weizhu Chen},
    booktitle={International Conference on Learning Representations},
    year={2022},
    url={https://openreview.net/forum?id=nZeVKeeFYf9}
}

@inproceedings{houlsby2019parameter,
  title={Parameter-efficient transfer learning for NLP},
  author={Houlsby, Neil and Giurgiu, Andrei and Jastrzebski, Stanislaw and Morrone, Bruna and De Laroussilhe, Quentin and Gesmundo, Andrea and Attariyan, Mona and Gelly, Sylvain},
  booktitle={International conference on machine learning},
  pages={2790--2799},
  year={2019},
  organization={PMLR}
}

@inproceedings{li2021prefix,
  title={Prefix-Tuning: Optimizing Continuous Prompts for Generation},
  author={Li, Xiang Lisa and Liang, Percy},
  booktitle={Proceedings of the 59th Annual Meeting of the Association for Computational Linguistics and the 11th International Joint Conference on Natural Language Processing (Volume 1: Long Papers)},
  pages={4582--4597},
  year={2021}
}

@inproceedings{lester2021power,
  title={The Power of Scale for Parameter-Efficient Prompt Tuning},
  author={Lester, Brian and Al-Rfou, Rami and Constant, Noah},
  booktitle={Proceedings of the 2021 Conference on Empirical Methods in Natural Language Processing},
  year={2021},
  organization={Association for Computational Linguistics}
}

@article{hansen2016cma,
  title={The CMA evolution strategy: A tutorial},
  author={Hansen, Nikolaus},
  journal={arXiv preprint arXiv:1604.00772},
  year={2016}
}

@inproceedings{
yu2024language,
title={Language Models are Super Mario: Absorbing Abilities from Homologous Models as a Free Lunch},
author={Le Yu and Bowen Yu and Haiyang Yu and Fei Huang and Yongbin Li},
booktitle={Forty-first International Conference on Machine Learning},
year={2024},
url={https://openreview.net/forum?id=fq0NaiU8Ex}
}

@misc{choi2024revisiting,
      title={Revisiting Weight Averaging for Model Merging}, 
      author={Jiho Choi and Donggyun Kim and Chanhyuk Lee and Seunghoon Hong},
      year={2025},
      eprint={2412.12153},
      archivePrefix={arXiv},
      primaryClass={cs.LG},
      url={https://arxiv.org/abs/2412.12153}, 
}

@Inproceedings{grattafiori2024llama,
 author = {Abhimanyu Dubey and Abhinav Jauhri and Abhinav Pandey and Abhishek Kadian and Ahmad Al-Dahle and Aiesha Letman and Akhil Mathur and A. Schelten and Amy Yang and Angela Fan and Anirudh Goyal and Anthony S. Hartshorn and Aobo Yang and Archi Mitra and A. Sravankumar and A. Korenev and Arthur Hinsvark and Arun Rao and Aston Zhang and Aur'elien Rodriguez and Austen Gregerson and Ava Spataru and Baptiste Rozière and Bethany Biron and Binh Tang and Bobbie Chern and C. Caucheteux and Chaya Nayak and Chloe Bi and Chris Marra and Chris McConnell and Christian Keller and C. Touret and Chunyang Wu and Corinne Wong and Cristian Canton Ferrer and Cyrus Nikolaidis and Damien Allonsius and Daniel Song and Danielle Pintz and Danny Livshits and David Esiobu and Dhruv Choudhary and Dhruv Mahajan and Diego Garcia-Olano and Diego Perino and Dieuwke Hupkes and Egor Lakomkin and Ehab A. AlBadawy and E. Lobanova and Emily Dinan and Eric Michael Smith and Filip Radenovic and Frank Zhang and Gabriele Synnaeve and Gabrielle Lee and Georgia Lewis Anderson and Graeme Nail and G. Mialon and Guanglong Pang and Guillem Cucurell and Hailey Nguyen and Hannah Korevaar and Hu Xu and Hugo Touvron and Iliyan Zarov and Imanol Arrieta Ibarra and Isabel M. Kloumann and Ishan Misra and Ivan Evtimov and Jade Copet and Jaewon Lee and Jan Geffert and Jana Vranes and Jason Park and Jay Mahadeokar and Jeet Shah and J. V. D. Linde and Jennifer Billock and Jenny Hong and Jenya Lee and J. Fu and Jianfeng Chi and Jianyu Huang and Jiawen Liu and Jie Wang and Jiecao Yu and Joanna Bitton and Joe Spisak and Jongsoo Park and Joseph Rocca and J. Johnstun and Joshua Saxe and Ju-Qing Jia and Kalyan Vasuden Alwala and K. Upasani and Kate Plawiak and Keqian Li and Kenneth Heafield and Kevin R. Stone and Khalid El-Arini and Krithika Iyer and Kshitiz Malik and Kuen-ley Chiu and Kunal Bhalla and Lauren Rantala-Yeary and L. Maaten and Lawrence Chen and Liang Tan and Liz Jenkins and Louis Martin and Lovish Madaan and Lubo Malo and Lukas Blecher and Lukas Landzaat and Luke de Oliveira and M. Muzzi and Ma-hesh Pasupuleti and Mannat Singh and Manohar Paluri and Marcin Kardas and Mathew Oldham and Mathieu Rita and Maya Pavlova and M. Kambadur and Mike Lewis and Min Si and Mitesh Kumar Singh and Mona Hassan and Naman Goyal and Narjes Torabi and Niko-lay Bashlykov and Nikolay Bogoychev and Niladri S. Chatterji and Olivier Duchenne and Onur cCelebi and Patrick Alrassy and Pengchuan Zhang and Pengwei Li and Petar Vasić and Peter Weng and Prajjwal Bhargava and P. Dubal and Praveen Krishnan and Punit Singh Koura and Puxin Xu and Qing He and Qingxiao Dong and Ragavan Srinivasan and Raj Ganapathy and Ramon Calderer and Ricardo Silveira Cabral and Robert Stojnic and R. Raileanu and Rohit Girdhar and Rohit Patel and Romain Sauvestre and Ron-nie Polidoro and Roshan Sumbaly and Ross Taylor and Ruan Silva and Rui Hou and Rui Wang and S. Hosseini and Sa-hana Chennabasappa and Sanjay Singh and Sean Bell and S. Kim and Sergey Edunov and Shaoliang Nie and Sharan Narang and S. Raparthy and Sheng Shen and Shengye Wan and Shruti Bhosale and Shun Zhang and Simon Vandenhende and Soumya Batra and Spencer Whitman and Sten Sootla and S. Collot and Suchin Gururangan and S. Borodinsky and Tamar Herman and Tara Fowler and Tarek Sheasha and Thomas Georgiou and Thomas Scialom and Tobias Speckbacher and Todor Mihaylov and Tong Xiao and Ujjwal Karn and Vedanuj Goswami and Vibhor Gupta and Vignesh Ramanathan and Viktor Kerkez and Vincent Gonguet and Vir-ginie Do and Vish Vogeti and Vladan Petrovic and Weiwei Chu and Wenhan Xiong and Wenyin Fu and Whit-ney Meers and Xavier Martinet and Xiaodong Wang and Xiaoqing Ellen Tan and Xinfeng Xie and Xuchao Jia and Xuewei Wang and Yaelle Goldschlag and Yashesh Gaur and Yasmine Babaei and Yiqian Wen and Yiwen Song and Yuchen Zhang and Yue Li and Yuning Mao and Zacharie Delpierre Coudert and Zhengxu Yan and Zhengxing Chen and Zoe Papakipos and Aaditya K. Singh and Aaron Grattafiori and Abha Jain and Adam Kelsey and Adam Shajnfeld and Adi Gangidi and Adolfo Victoria and Ahuva Goldstand and A. Menon and Ajay Sharma and Alex Boesenberg and Alex Vaughan and Alexei Baevski and Allie Feinstein and Amanda Kallet and Amit Sangani and Anam Yunus and Andrei Lupu and Andres Alvarado and A. Caples and Andrew Gu and Andrew Ho and Andrew Poulton and Andrew Ryan and Ankit Ramchandani and Annie Franco and Aparajita Saraf and Arkabandhu Chowdhury and Ashley Gabriel and Ashwin Bharambe and Assaf Eisenman and Azadeh Yazdan and Beau James and Ben Maurer and B. Leonhardi and Po-Yao (Bernie) Huang and Beth Loyd and Beto de Paola and Bhargavi Paranjape and Bing Liu and Bo Wu and B. Ni and Braden Hancock and Bram Wasti and Brandon Spence and B. Stojkovic and Brian Gamido and Britt Montalvo and Carl Parker and Carly Burton and Catalina Mejia and Changhan Wang and Changkyu Kim and Chao Zhou and Chester Hu and Ching-Hsiang Chu and Chris Cai and Chris Tindal and Christoph Feichtenhofer and Damon Civin and Dana Beaty and Daniel Kreymer and Shang-Wen Li and Danny Wyatt and David Adkins and David Xu and Davide Testuggine and Delia David and Devi Parikh and Diana Liskovich and Didem Foss and Dingkang Wang and Duc Le and Dustin Holland and Edward Dowling and Eissa Jamil and Elaine Montgomery and Eleonora Presani and Emily Hahn and Emily Wood and Erik Brinkman and E. Arcaute and Evan Dunbar and Evan Smothers and Fei Sun and F. Kreuk and Feng Tian and Firat Ozgenel and Francesco Caggioni and Francisco (Paco) Guzmán and Frank J. Kanayet and Frank Seide and Gabriela Medina Florez and Gabriella Schwarz and Gada Badeer and Georgia Swee and Gil Halpern and Govind Thattai and Grant Herman and G. Sizov and Guangyi Zhang and G. Lakshminarayanan and Hamid Shojanazeri and Han Zou and Hannah Wang and Han Zha and Haroun Habeeb and Harrison Rudolph and Helen Suk and Henry Aspegren and Hunter Goldman and Igor Molybog and Igor Tufanov and Irina-Elena Veliche and Itai Gat and Jake Weissman and James Geboski and James Kohli and Japhet Asher and Jean-Baptiste Gaya and Jeff Marcus and Jeff Tang and Jennifer Chan and Jenny Zhen and J. Reizenstein and J. Teboul and Jessica Zhong and Jian Jin and Jingyi Yang and Joe Cummings and Jon Carvill and Jon Shepard and J. McPhie and Jonathan Torres and Josh Ginsburg and Junjie Wang and Kaixing(Kai) Wu and U. KamHou and Karan Saxena and Karthik Prasad and Kartikay Khandelwal and Katayoun Zand and Kathy Matosich and K. Veeraraghavan and Kelly Michelena and Keqian Li and Kun Huang and Kunal Chawla and Kushal Lakhotia and Kyle Huang and Lailin Chen and Lakshya Garg and A. Lavender and Leandro Silva and Lee Bell and Lei Zhang and Liangpeng Guo and Licheng Yu and Liron Moshkovich and Luca Wehrstedt and Madian Khabsa and Manav Avalani and M. Bhatt and M. Tsimpoukelli and Martynas Mankus and Matan Hasson and Matthias Lennie and Matthias Reso and Maxim Groshev and Maxim Naumov and Maya Lathi and Meghan Keneally and M. Seltzer and Michal Valko and Michelle Restrepo and Mihir Patel and Mik Vyatskov and Mikayel Samvelyan and Mike Clark and M. Macey and Mike Wang and Miquel Jubert Hermoso and Mo Metanat and Mohammad Rastegari and Mun-ish Bansal and N. Santhanam and Natascha Parks and Natasha White and Navy-ata Bawa and Nayan Singhal and Nick Egebo and Nicolas Usunier and Nikolay Pavlovich Laptev and Ning Dong and Ning Zhang and Norman Cheng and Oleg Chernoguz and O. Hart and Omkar Salpekar and Ozlem Kalinli and Parkin Kent and Parth Parekh and Paul Saab and Pavan Balaji and Pe-dro Rittner and Philip Bontrager and Pierre Roux and Piotr Dollár and Polina Zvyagina and Prashant Ratanchandani and Pritish Yuvraj and Qian Liang and Rachad Alao and Rachel Rodriguez and Rafi Ayub and Raghotham Murthy and Raghu Nayani and Rahul Mitra and Raymond Li and Rebekkah Hogan and Robin Battey and Rocky Wang and R. Maheswari and Russ Howes and Ruty Rinott and Sai Jayesh Bondu and Samyak Datta and Sara Chugh and Sara Hunt and Sargun Dhillon and S. Sidorov and Satadru Pan and Saurabh Verma and Seiji Yamamoto and Sharadh Ramaswamy and Shaun Lindsay and Sheng Feng and Shenghao Lin and S. Zha and Shiva Shankar and Shuqiang Zhang and Sinong Wang and Sneha Agarwal and S. Sajuyigbe and Soumith Chintala and Stephanie Max and Stephen Chen and Steve Kehoe and Steve Satterfield and Sudarshan Govindaprasad and S. Gupta and Sung-Bae Cho and Sunny Virk and Suraj Subramanian and Sy Choudhury and Sydney Goldman and Tal Remez and Tamar Glaser and Tamara Best and Thilo Kohler and Thomas Robinson and Tianhe Li and Tianjun Zhang and Tim Matthews and Timothy Chou and Tzook Shaked and Varun Vontimitta and Victoria Ajayi and Victoria Montanez and Vijai Mohan and Vinay Satish Kumar and Vishal Mangla and Vlad Ionescu and V. Poenaru and Vlad T. Mihailescu and Vladimir Ivanov and Wei Li and Wenchen Wang and Wenwen Jiang and Wes Bouaziz and Will Constable and Xia Tang and Xiaofang Wang and Xiaojian Wu and Xiaolan Wang and Xide Xia and Xilun Wu and Xinbo Gao and Yanjun Chen and Ye Hu and Ye Jia and Ye Qi and Yenda Li and Yilin Zhang and Ying Zhang and Yossi Adi and Youngjin Nam and Yu Wang and Yuchen Hao and Yundi Qian and Yuzi He and Zach Rait and Zachary DeVito and Zef Rosnbrick and Zhaoduo Wen and Zhenyu Yang and Zhiwei Zhao},
 title = {The Llama 3 Herd of Models},
 year = {2024}
}

@misc{yang2025qwen3,
      title={Qwen3 Technical Report}, 
      author={An Yang and Anfeng Li and Baosong Yang and Beichen Zhang and Binyuan Hui and Bo Zheng and Bowen Yu and Chang Gao and Chengen Huang and Chenxu Lv and Chujie Zheng and Dayiheng Liu and Fan Zhou and Fei Huang and Feng Hu and Hao Ge and Haoran Wei and Huan Lin and Jialong Tang and Jian Yang and Jianhong Tu and Jianwei Zhang and Jianxin Yang and Jiaxi Yang and Jing Zhou and Jingren Zhou and Junyang Lin and Kai Dang and Keqin Bao and Kexin Yang and Le Yu and Lianghao Deng and Mei Li and Mingfeng Xue and Mingze Li and Pei Zhang and Peng Wang and Qin Zhu and Rui Men and Ruize Gao and Shixuan Liu and Shuang Luo and Tianhao Li and Tianyi Tang and Wenbiao Yin and Xingzhang Ren and Xinyu Wang and Xinyu Zhang and Xuancheng Ren and Yang Fan and Yang Su and Yichang Zhang and Yinger Zhang and Yu Wan and Yuqiong Liu and Zekun Wang and Zeyu Cui and Zhenru Zhang and Zhipeng Zhou and Zihan Qiu},
      year={2025},
      eprint={2505.09388},
      archivePrefix={arXiv},
      primaryClass={cs.CL},
      url={https://arxiv.org/abs/2505.09388}, 
}

@misc{achiam2023gpt,
      title={{GPT-4} Technical Report}, 
      author={OpenAI and Josh Achiam and Steven Adler and Sandhini Agarwal and Lama Ahmad and Ilge Akkaya and Florencia Leoni Aleman and Diogo Almeida and Janko Altenschmidt and Sam Altman and Shyamal Anadkat and Red Avila and Igor Babuschkin and Suchir Balaji and Valerie Balcom and Paul Baltescu and Haiming Bao and Mohammad Bavarian and Jeff Belgum and Irwan Bello and Jake Berdine and Gabriel Bernadett-Shapiro and Christopher Berner and Lenny Bogdonoff and Oleg Boiko and Madelaine Boyd and Anna-Luisa Brakman and Greg Brockman and Tim Brooks and Miles Brundage and Kevin Button and Trevor Cai and Rosie Campbell and Andrew Cann and Brittany Carey and Chelsea Carlson and Rory Carmichael and Brooke Chan and Che Chang and Fotis Chantzis and Derek Chen and Sully Chen and Ruby Chen and Jason Chen and Mark Chen and Ben Chess and Chester Cho and Casey Chu and Hyung Won Chung and Dave Cummings and Jeremiah Currier and Yunxing Dai and Cory Decareaux and Thomas Degry and Noah Deutsch and Damien Deville and Arka Dhar and David Dohan and Steve Dowling and Sheila Dunning and Adrien Ecoffet and Atty Eleti and Tyna Eloundou and David Farhi and Liam Fedus and Niko Felix and Simón Posada Fishman and Juston Forte and Isabella Fulford and Leo Gao and Elie Georges and Christian Gibson and Vik Goel and Tarun Gogineni and Gabriel Goh and Rapha Gontijo-Lopes and Jonathan Gordon and Morgan Grafstein and Scott Gray and Ryan Greene and Joshua Gross and Shixiang Shane Gu and Yufei Guo and Chris Hallacy and Jesse Han and Jeff Harris and Yuchen He and Mike Heaton and Johannes Heidecke and Chris Hesse and Alan Hickey and Wade Hickey and Peter Hoeschele and Brandon Houghton and Kenny Hsu and Shengli Hu and Xin Hu and Joost Huizinga and Shantanu Jain and Shawn Jain and Joanne Jang and Angela Jiang and Roger Jiang and Haozhun Jin and Denny Jin and Shino Jomoto and Billie Jonn and Heewoo Jun and Tomer Kaftan and Łukasz Kaiser and Ali Kamali and Ingmar Kanitscheider and Nitish Shirish Keskar and Tabarak Khan and Logan Kilpatrick and Jong Wook Kim and Christina Kim and Yongjik Kim and Jan Hendrik Kirchner and Jamie Kiros and Matt Knight and Daniel Kokotajlo and Łukasz Kondraciuk and Andrew Kondrich and Aris Konstantinidis and Kyle Kosic and Gretchen Krueger and Vishal Kuo and Michael Lampe and Ikai Lan and Teddy Lee and Jan Leike and Jade Leung and Daniel Levy and Chak Ming Li and Rachel Lim and Molly Lin and Stephanie Lin and Mateusz Litwin and Theresa Lopez and Ryan Lowe and Patricia Lue and Anna Makanju and Kim Malfacini and Sam Manning and Todor Markov and Yaniv Markovski and Bianca Martin and Katie Mayer and Andrew Mayne and Bob McGrew and Scott Mayer McKinney and Christine McLeavey and Paul McMillan and Jake McNeil and David Medina and Aalok Mehta and Jacob Menick and Luke Metz and Andrey Mishchenko and Pamela Mishkin and Vinnie Monaco and Evan Morikawa and Daniel Mossing and Tong Mu and Mira Murati and Oleg Murk and David Mély and Ashvin Nair and Reiichiro Nakano and Rajeev Nayak and Arvind Neelakantan and Richard Ngo and Hyeonwoo Noh and Long Ouyang and Cullen O'Keefe and Jakub Pachocki and Alex Paino and Joe Palermo and Ashley Pantuliano and Giambattista Parascandolo and Joel Parish and Emy Parparita and Alex Passos and Mikhail Pavlov and Andrew Peng and Adam Perelman and Filipe de Avila Belbute Peres and Michael Petrov and Henrique Ponde de Oliveira Pinto and Michael and Pokorny and Michelle Pokrass and Vitchyr H. Pong and Tolly Powell and Alethea Power and Boris Power and Elizabeth Proehl and Raul Puri and Alec Radford and Jack Rae and Aditya Ramesh and Cameron Raymond and Francis Real and Kendra Rimbach and Carl Ross and Bob Rotsted and Henri Roussez and Nick Ryder and Mario Saltarelli and Ted Sanders and Shibani Santurkar and Girish Sastry and Heather Schmidt and David Schnurr and John Schulman and Daniel Selsam and Kyla Sheppard and Toki Sherbakov and Jessica Shieh and Sarah Shoker and Pranav Shyam and Szymon Sidor and Eric Sigler and Maddie Simens and Jordan Sitkin and Katarina Slama and Ian Sohl and Benjamin Sokolowsky and Yang Song and Natalie Staudacher and Felipe Petroski Such and Natalie Summers and Ilya Sutskever and Jie Tang and Nikolas Tezak and Madeleine B. Thompson and Phil Tillet and Amin Tootoonchian and Elizabeth Tseng and Preston Tuggle and Nick Turley and Jerry Tworek and Juan Felipe Cerón Uribe and Andrea Vallone and Arun Vijayvergiya and Chelsea Voss and Carroll Wainwright and Justin Jay Wang and Alvin Wang and Ben Wang and Jonathan Ward and Jason Wei and CJ Weinmann and Akila Welihinda and Peter Welinder and Jiayi Weng and Lilian Weng and Matt Wiethoff and Dave Willner and Clemens Winter and Samuel Wolrich and Hannah Wong and Lauren Workman and Sherwin Wu and Jeff Wu and Michael Wu and Kai Xiao and Tao Xu and Sarah Yoo and Kevin Yu and Qiming Yuan and Wojciech Zaremba and Rowan Zellers and Chong Zhang and Marvin Zhang and Shengjia Zhao and Tianhao Zheng and Juntang Zhuang and William Zhuk and Barret Zoph},
      year={2024},
      eprint={2303.08774},
      archivePrefix={arXiv},
      primaryClass={cs.CL},
      url={https://arxiv.org/abs/2303.08774}, 
}

@article{dettmers2023qlora,
  title={Qlora: Efficient finetuning of quantized llms},
  author={Dettmers, Tim and Pagnoni, Artidoro and Holtzman, Ari and Zettlemoyer, Luke},
  journal={Advances in neural information processing systems},
  volume={36},
  pages={10088--10115},
  year={2023}
}

@inproceedings{
    zhang2023adaptive,
    title={Adaptive Budget Allocation for Parameter-Efficient Fine-Tuning },
    author={Qingru Zhang and Minshuo Chen and Alexander Bukharin and Pengcheng He and Yu Cheng and Weizhu Chen and Tuo Zhao},
    booktitle={The Eleventh International Conference on Learning Representations },
    year={2023},
    url={https://openreview.net/forum?id=lq62uWRJjiY}
}

@inproceedings{
wei2022finetuned,
title={Finetuned Language Models are Zero-Shot Learners},
author={Jason Wei and Maarten Bosma and Vincent Zhao and Kelvin Guu and Adams Wei Yu and Brian Lester and Nan Du and Andrew M. Dai and Quoc V Le},
booktitle={International Conference on Learning Representations},
year={2022},
url={https://openreview.net/forum?id=gEZrGCozdqR}
}

@inproceedings{
sanh2022multitask,
title={Multitask Prompted Training Enables Zero-Shot Task Generalization},
author={Victor Sanh and Albert Webson and Colin Raffel and Stephen Bach and Lintang Sutawika and Zaid Alyafeai and Antoine Chaffin and Arnaud Stiegler and Arun Raja and Manan Dey and M Saiful Bari and Canwen Xu and Urmish Thakker and Shanya Sharma Sharma and Eliza Szczechla and Taewoon Kim and Gunjan Chhablani and Nihal Nayak and Debajyoti Datta and Jonathan Chang and Mike Tian-Jian Jiang and Han Wang and Matteo Manica and Sheng Shen and Zheng Xin Yong and Harshit Pandey and Rachel Bawden and Thomas Wang and Trishala Neeraj and Jos Rozen and Abheesht Sharma and Andrea Santilli and Thibault Fevry and Jason Alan Fries and Ryan Teehan and Teven Le Scao and Stella Biderman and Leo Gao and Thomas Wolf and Alexander M Rush},
booktitle={International Conference on Learning Representations},
year={2022},
url={https://openreview.net/forum?id=9Vrb9D0WI4}
}

@InProceedings{radford2021learning,
  title = 	 {Learning Transferable Visual Models From Natural Language Supervision},
  author =       {Radford, Alec and Kim, Jong Wook and Hallacy, Chris and Ramesh, Aditya and Goh, Gabriel and Agarwal, Sandhini and Sastry, Girish and Askell, Amanda and Mishkin, Pamela and Clark, Jack and Krueger, Gretchen and Sutskever, Ilya},
  booktitle = 	 {Proceedings of the 38th International Conference on Machine Learning},
  pages = 	 {8748--8763},
  year = 	 {2021},
  editor = 	 {Meila, Marina and Zhang, Tong},
  volume = 	 {139},
  series = 	 {Proceedings of Machine Learning Research},
  month = 	 {18--24 Jul},
  publisher =    {PMLR},
  url = 	 {https://proceedings.mlr.press/v139/radford21a.html}
}

@inproceedings{bowman2015large,
  title={A large annotated corpus for learning natural language inference},
  author={Bowman, Samuel and Angeli, Gabor and Potts, Christopher and Manning, Christopher D},
  booktitle={Proceedings of the 2015 conference on empirical methods in natural language processing},
  pages={632--642},
  year={2015}
}

@inproceedings{williams2018broad,
  title={A broad-coverage challenge corpus for sentence understanding through inference},
  author={Williams, Adina and Nangia, Nikita and Bowman, Samuel},
  booktitle={Proceedings of the 2018 conference of the North American chapter of the association for computational linguistics: human language technologies, volume 1 (long papers)},
  pages={1112--1122},
  year={2018}
}

@inproceedings{marelli-etal-2014-sick,
    title = "A {SICK} cure for the evaluation of compositional distributional semantic models",
    author = "Marelli, Marco  and
      Menini, Stefano  and
      Baroni, Marco  and
      Bentivogli, Luisa  and
      Bernardi, Raffaella  and
      Zamparelli, Roberto",
    editor = "Calzolari, Nicoletta  and
      Choukri, Khalid  and
      Declerck, Thierry  and
      Loftsson, Hrafn  and
      Maegaard, Bente  and
      Mariani, Joseph  and
      Moreno, Asuncion  and
      Odijk, Jan  and
      Piperidis, Stelios",
    booktitle = "Proceedings of the Ninth International Conference on Language Resources and Evaluation ({LREC}'14)",
    month = may,
    year = "2014",
    address = "Reykjavik, Iceland",
    publisher = "European Language Resources Association (ELRA)",
    url = "https://aclanthology.org/L14-1314/",
    pages = "216--223"
}

@inproceedings{wang-etal-2018-glue,
    title = "{GLUE}: A Multi-Task Benchmark and Analysis Platform for Natural Language Understanding",
    author = "Wang, Alex  and
      Singh, Amanpreet  and
      Michael, Julian  and
      Hill, Felix  and
      Levy, Omer  and
      Bowman, Samuel",
    editor = "Linzen, Tal  and
      Chrupa{\l}a, Grzegorz  and
      Alishahi, Afra",
    booktitle = "Proceedings of the 2018 {EMNLP} Workshop {B}lackbox{NLP}: Analyzing and Interpreting Neural Networks for {NLP}",
    month = nov,
    year = "2018",
    address = "Brussels, Belgium",
    publisher = "Association for Computational Linguistics",
    url = "https://aclanthology.org/W18-5446/",
    doi = "10.18653/v1/W18-5446",
    pages = "353--355"
}

@inproceedings{khot2018scitail,
  title={Scitail: A textual entailment dataset from science question answering},
  author={Khot, Tushar and Sabharwal, Ashish and Clark, Peter},
  booktitle={Proceedings of the AAAI conference on artificial intelligence},
  volume={32},
  number={1},
  year={2018}
}

@inproceedings{krause20133d,
  title={3d object representations for fine-grained categorization},
  author={Krause, Jonathan and Stark, Michael and Deng, Jia and Fei-Fei, Li},
  booktitle={Proceedings of the IEEE international conference on computer vision workshops},
  pages={554--561},
  year={2013}
}

@inproceedings{cimpoi2014describing,
  title={Describing textures in the wild},
  author={Cimpoi, Mircea and Maji, Subhransu and Kokkinos, Iasonas and Mohamed, Sammy and Vedaldi, Andrea},
  booktitle={Proceedings of the IEEE conference on computer vision and pattern recognition},
  pages={3606--3613},
  year={2014}
}

@article{helber2019eurosat,
  title={Eurosat: A novel dataset and deep learning benchmark for land use and land cover classification},
  author={Helber, Patrick and Bischke, Benjamin and Dengel, Andreas and Borth, Damian},
  journal={IEEE Journal of Selected Topics in Applied Earth Observations and Remote Sensing},
  volume={12},
  number={7},
  pages={2217--2226},
  year={2019},
  publisher={IEEE}
}

@inproceedings{stallkamp2011german,
  title={The German traffic sign recognition benchmark: a multi-class classification competition},
  author={Stallkamp, Johannes and Schlipsing, Marc and Salmen, Jan and Igel, Christian},
  booktitle={The 2011 international joint conference on neural networks},
  pages={1453--1460},
  year={2011},
  organization={IEEE}
}

@article{cheng2017remote,
  title={Remote sensing image scene classification: Benchmark and state of the art},
  author={Cheng, Gong and Han, Junwei and Lu, Xiaoqiang},
  journal={Proceedings of the IEEE},
  volume={105},
  number={10},
  pages={1865--1883},
  year={2017},
  publisher={IEEE}
}

@article{xiao2016sun,
  title={Sun database: Exploring a large collection of scene categories},
  author={Xiao, Jianxiong and Ehinger, Krista A and Hays, James and Torralba, Antonio and Oliva, Aude},
  journal={International Journal of Computer Vision},
  volume={119},
  number={1},
  pages={3--22},
  year={2016},
  publisher={Springer}
}

@inproceedings{netzer2011reading,
  title={Reading Digits in Natural Images with Unsupervised Feature Learning},
  author={Yuval Netzer and Tao Wang and Adam Coates and A. Bissacco and Bo Wu and A. Ng},
  year={2011},
  url={https://api.semanticscholar.org/CorpusID:16852518}
}

@article{lecun-mnisthandwrittendigit-2010,
  author = {LeCun, Yann and Cortes, Corinna},
  howpublished = {http://yann.lecun.com/exdb/mnist/},
  lastchecked = {2016-01-14 14:24:11},
  title = {{MNIST} handwritten digit database},
  url = {http://yann.lecun.com/exdb/mnist/},
  username = {mhwombat},
  year = 2010
}

\appendix


\section{Evolutionary Search Details}
\label{app:evolutionary_search}

In this section, we provide the implementation details of the evolutionary search strategy employed in our framework. Algorithm~\ref{alg:enmp} outlines the complete optimization process, including the initialization of the CMA-ES strategy, the generation of candidate pruning masks via latent variable mapping, and the iterative update mechanism driven by validation performance.

\begin{algorithm}[htbp!]
\small
\caption{Evolutionary Search with CMA-ES}
\label{alg:enmp}
\begin{algorithmic}[1]
\REQUIRE Pre-trained LoRA experts $\mathcal{T} = \{\Delta \mathbf{W}_1, \dots, \Delta \mathbf{W}_T\}$, Validation Set $\mathcal{D}_\text{val}$, Generations $G$, Pop. Size $N_\text{pop}$.
\STATE \textbf{Initialize:} Mean $\boldsymbol{\mu}^{(0)} \leftarrow -\mathbf{1}$, Covariance $\mathbf{C}^{(0)} \leftarrow \mathbf{I}$, Step size $\sigma$.
\FOR{generation $g = 0$ \TO $G-1$}
    \STATE Sample $N_\text{pop}$ latent vectors: 
    \STATE \quad $\mathbf{z}_i \sim \mathcal{N}(\boldsymbol{\mu}^{(g)}, (\sigma^{(g)})^2 \mathbf{C}^{(g)})$
    \FOR{$i = 1$ \TO $N_\text{pop}$}
        \STATE \textbf{Masking:} $\mathbf{m}_i^\text{flat} \leftarrow \textsc{TopK-Map}(\mathbf{z}_i)$ \hfill \textit{\footnotesize(Eq.~\ref{eq:mapping})}
        \STATE \textbf{Reshape:} $\mathbf{m}_i \leftarrow \mathbf{m}_i^\text{flat}$
        \STATE \textbf{Pruning:} $\mathcal{S}_i \leftarrow \{ \Delta \mathbf{W}_t^{(l)} \mid m_{i, l,t} = 0 \}$
        \STATE \textbf{Merging:} $\mathbf{W}_{\text{merged}} \leftarrow \Phi(\mathcal{S}_i)$ \hfill \textit{\footnotesize(Eq.~\ref{eq:sta_merge})}
        \STATE \textbf{Eval:} $\mathcal{F}_i \leftarrow \mathcal{M}(\mathbf{W}_{\text{merged}}; \mathcal{D}_\text{val})$
        \STATE Update best mask $\mathbf{m}^*$ if $\mathcal{F}_i > \mathcal{F}_{best}$.
    \ENDFOR
    \STATE Update $\boldsymbol{\mu}^{(g+1)}, \mathbf{C}^{(g+1)}$ via CMA-ES rule using $\{\mathcal{F}_i\}$.
\ENDFOR
\STATE \textbf{Output:} Optimal merged model $\mathbf{W}_{\text{final}}$ using $\mathbf{m}^*$.
\end{algorithmic}
\end{algorithm}

\begin{table}[htbp!]
    \centering
    \scriptsize 
    
    \resizebox{\linewidth}{!}{
    \begin{tabular}{l cccccc}
    \toprule
    \textbf{Layer} & \textbf{SNLI} & \textbf{MNLI} & \textbf{SICK} & \textbf{QNLI} & \textbf{RTE} & \textbf{SCITAIL} \\
    \midrule
    0 & \cellcolor[RGB]{236,157,162}+0.5025 & \cellcolor[RGB]{229,119,126}+0.6965 & \cellcolor[RGB]{253,254,254}-0.0165 & \cellcolor[RGB]{254,250,250}+0.0249 & \cellcolor[RGB]{252,240,241}+0.0759 & \cellcolor[RGB]{247,216,218}+0.1954 \\
    1 & \cellcolor[RGB]{246,211,213}+0.2244 & \cellcolor[RGB]{239,171,176}+0.4301 & \cellcolor[RGB]{254,253,253}+0.0106 & \cellcolor[RGB]{249,223,225}+0.1585 & \cellcolor[RGB]{247,217,219}+0.1909 & \cellcolor[RGB]{249,225,227}+0.1493 \\
    2 & \cellcolor[RGB]{248,221,223}+0.1694 & \cellcolor[RGB]{244,199,202}+0.2833 & \cellcolor[RGB]{254,250,250}+0.0237 & \cellcolor[RGB]{249,228,230}+0.1357 & \cellcolor[RGB]{253,246,247}+0.0424 & \cellcolor[RGB]{248,219,221}+0.1808 \\
    3 & \cellcolor[RGB]{253,247,248}+0.0374 & \cellcolor[RGB]{253,245,246}+0.0464 & \cellcolor[RGB]{254,254,254}+0.0007 & \cellcolor[RGB]{250,251,253}-0.0441 & \cellcolor[RGB]{254,252,252}+0.0121 & \cellcolor[RGB]{254,250,250}+0.0233 \\
    4 & \cellcolor[RGB]{248,250,252}-0.0602 & \cellcolor[RGB]{252,242,243}+0.0643 & \cellcolor[RGB]{250,232,234}+0.1127 & \cellcolor[RGB]{250,251,252}-0.0450 & \cellcolor[RGB]{254,252,252}+0.0152 & \cellcolor[RGB]{249,226,228}+0.1436 \\
    5 & \cellcolor[RGB]{246,209,211}+0.2331 & \cellcolor[RGB]{248,222,224}+0.1672 & \cellcolor[RGB]{248,221,223}+0.1728 & \cellcolor[RGB]{240,244,247}-0.1481 & \cellcolor[RGB]{254,254,254}+0.0042 & \cellcolor[RGB]{252,242,243}+0.0642 \\
    6 & \cellcolor[RGB]{251,233,235}+0.1073 & \cellcolor[RGB]{246,208,210}+0.2377 & \cellcolor[RGB]{252,243,244}+0.0580 & \cellcolor[RGB]{240,244,247}-0.1480 & \cellcolor[RGB]{254,250,250}+0.0220 & \cellcolor[RGB]{254,251,251}+0.0203 \\
    7 & \cellcolor[RGB]{253,246,247}+0.0433 & \cellcolor[RGB]{250,233,234}+0.1108 & \cellcolor[RGB]{250,231,233}+0.1214 & \cellcolor[RGB]{244,247,249}-0.1064 & \cellcolor[RGB]{248,250,251}-0.0652 & \cellcolor[RGB]{252,243,244}+0.0577 \\
    8 & \cellcolor[RGB]{252,242,243}+0.0660 & \cellcolor[RGB]{236,158,162}+0.4925 & \cellcolor[RGB]{249,227,229}+0.1411 & \cellcolor[RGB]{227,235,241}-0.2709 & \cellcolor[RGB]{248,218,220}+0.1860 & \cellcolor[RGB]{252,242,243}+0.0645 \\
    9 & \cellcolor[RGB]{252,243,244}+0.0596 & \cellcolor[RGB]{249,224,226}+0.1534 & \cellcolor[RGB]{247,216,218}+0.1944 & \cellcolor[RGB]{227,235,241}-0.2721 & \cellcolor[RGB]{249,223,225}+0.1591 & \cellcolor[RGB]{253,246,247}+0.0412 \\
    10 & \cellcolor[RGB]{252,239,240}+0.0809 & \cellcolor[RGB]{247,216,218}+0.1962 & \cellcolor[RGB]{249,226,227}+0.1464 & \cellcolor[RGB]{243,246,249}-0.1141 & \cellcolor[RGB]{249,227,228}+0.1425 & \cellcolor[RGB]{254,254,254}+0.0013 \\
    11 & \cellcolor[RGB]{239,172,177}+0.4212 & \cellcolor[RGB]{245,206,209}+0.2474 & \cellcolor[RGB]{245,202,205}+0.2655 & \cellcolor[RGB]{206,220,231}-0.4792 & \cellcolor[RGB]{244,199,202}+0.2823 & \cellcolor[RGB]{236,158,162}+0.4918 \\
    12 & \cellcolor[RGB]{235,152,157}+0.5209 & \cellcolor[RGB]{228,113,121}+0.7198 & \cellcolor[RGB]{244,201,203}+0.2745 & \cellcolor[RGB]{154,183,205}-1.2861 & \cellcolor[RGB]{229,118,125}+0.6948 & \cellcolor[RGB]{250,229,230}+0.1298 \\
    13 & \cellcolor[RGB]{225,98,107}+0.7976 & \cellcolor[RGB]{218,59,70}+1.5032 & \cellcolor[RGB]{247,215,217}+0.2018 & \cellcolor[RGB]{214,226,235}-0.3989 & \cellcolor[RGB]{230,124,131}+0.6677 & \cellcolor[RGB]{250,232,234}+0.1148 \\
    14 & \cellcolor[RGB]{239,171,175}+0.4281 & \cellcolor[RGB]{244,197,200}+0.2933 & \cellcolor[RGB]{247,214,216}+0.2083 & \cellcolor[RGB]{225,233,240}-0.2945 & \cellcolor[RGB]{225,98,107}+0.7995 & \cellcolor[RGB]{252,239,240}+0.0798 \\
    15 & \cellcolor[RGB]{237,160,164}+0.4847 & \cellcolor[RGB]{218,59,70}+0.9975 & \cellcolor[RGB]{247,216,218}+0.1951 & \cellcolor[RGB]{214,226,235}-0.4015 & \cellcolor[RGB]{218,59,70}+1.1883 & \cellcolor[RGB]{251,238,239}+0.0847 \\
    16 & \cellcolor[RGB]{250,232,234}+0.1145 & \cellcolor[RGB]{239,171,176}+0.4264 & \cellcolor[RGB]{248,219,221}+0.1804 & \cellcolor[RGB]{190,208,222}-0.6437 & \cellcolor[RGB]{231,128,135}+0.6445 & \cellcolor[RGB]{251,235,237}+0.0982 \\
    17 & \cellcolor[RGB]{245,205,208}+0.2521 & \cellcolor[RGB]{250,227,229}+0.1381 & \cellcolor[RGB]{249,223,225}+0.1607 & \cellcolor[RGB]{230,237,242}-0.2474 & \cellcolor[RGB]{251,236,237}+0.0949 & \cellcolor[RGB]{249,227,229}+0.1409 \\
    18 & \cellcolor[RGB]{254,251,251}+0.0197 & \cellcolor[RGB]{249,224,226}+0.1548 & \cellcolor[RGB]{249,226,228}+0.1434 & \cellcolor[RGB]{254,254,254}+0.0318 & \cellcolor[RGB]{249,228,229}+0.1374 & \cellcolor[RGB]{248,222,224}+0.1658 \\
    19 & \cellcolor[RGB]{247,215,217}+0.1997 & \cellcolor[RGB]{253,245,246}+0.0495 & \cellcolor[RGB]{247,214,216}+0.2051 & \cellcolor[RGB]{254,252,253}+0.0114 & \cellcolor[RGB]{250,231,232}+0.1211 & \cellcolor[RGB]{249,225,227}+0.1487 \\
    20 & \cellcolor[RGB]{247,216,219}+0.1941 & \cellcolor[RGB]{247,214,217}+0.2056 & \cellcolor[RGB]{248,219,221}+0.1798 & \cellcolor[RGB]{251,239,240}+0.0813 & \cellcolor[RGB]{253,248,248}+0.0346 & \cellcolor[RGB]{253,246,247}+0.0439 \\
    21 & \cellcolor[RGB]{252,241,242}+0.0687 & \cellcolor[RGB]{251,252,253}-0.0392 & \cellcolor[RGB]{251,252,253}-0.0301 & \cellcolor[RGB]{246,248,250}-0.0865 & \cellcolor[RGB]{249,225,226}+0.1521 & \cellcolor[RGB]{251,234,235}+0.1051 \\
    22 & \cellcolor[RGB]{254,252,252}+0.0138 & \cellcolor[RGB]{241,245,248}-0.1376 & \cellcolor[RGB]{246,211,213}+0.2211 & \cellcolor[RGB]{249,250,252}-0.0590 & \cellcolor[RGB]{250,229,230}+0.1302 & \cellcolor[RGB]{254,250,251}+0.0217 \\
    23 & \cellcolor[RGB]{249,224,225}+0.1581 & \cellcolor[RGB]{249,224,226}+0.1541 & \cellcolor[RGB]{247,214,216}+0.2085 & \cellcolor[RGB]{249,227,228}+0.1425 & \cellcolor[RGB]{248,222,224}+0.1657 & \cellcolor[RGB]{248,220,222}+0.1765 \\
    24 & \cellcolor[RGB]{246,211,213}+0.2243 & \cellcolor[RGB]{254,254,254}-0.0065 & \cellcolor[RGB]{247,214,216}+0.2044 & \cellcolor[RGB]{254,250,251}+0.0209 & \cellcolor[RGB]{247,214,216}+0.2088 & \cellcolor[RGB]{248,222,224}+0.1654 \\
    25 & \cellcolor[RGB]{254,250,250}+0.0222 & \cellcolor[RGB]{248,222,224}+0.1643 & \cellcolor[RGB]{249,223,225}+0.1607 & \cellcolor[RGB]{254,250,250}+0.0249 & \cellcolor[RGB]{249,226,227}+0.1457 & \cellcolor[RGB]{252,244,244}+0.0544 \\
    26 & \cellcolor[RGB]{254,253,253}+0.0082 & \cellcolor[RGB]{254,252,252}+0.0137 & \cellcolor[RGB]{254,254,254}-0.0043 & \cellcolor[RGB]{255,255,255}+0.0005 & \cellcolor[RGB]{247,214,216}+0.2046 & \cellcolor[RGB]{253,254,254}-0.0121 \\
    27 & \cellcolor[RGB]{244,199,202}+0.2817 & \cellcolor[RGB]{247,213,215}+0.2123 & \cellcolor[RGB]{250,231,233}+0.1193 & \cellcolor[RGB]{241,185,189}+0.3553 & \cellcolor[RGB]{249,223,225}+0.1595 & \cellcolor[RGB]{248,222,224}+0.1639 \\
    28 & \cellcolor[RGB]{246,210,212}+0.2290 & \cellcolor[RGB]{251,236,237}+0.0965 & \cellcolor[RGB]{248,222,224}+0.1664 & \cellcolor[RGB]{252,242,243}+0.0615 & \cellcolor[RGB]{247,217,219}+0.1908 & \cellcolor[RGB]{249,225,226}+0.1506 \\
    29 & \cellcolor[RGB]{250,251,252}-0.0474 & \cellcolor[RGB]{237,243,246}-0.1692 & \cellcolor[RGB]{249,225,226}+0.1511 & \cellcolor[RGB]{252,240,241}+0.0747 & \cellcolor[RGB]{248,220,222}+0.1752 & \cellcolor[RGB]{250,251,252}-0.0420 \\
    30 & \cellcolor[RGB]{251,239,240}+0.0793 & \cellcolor[RGB]{234,147,152}+0.5465 & \cellcolor[RGB]{249,227,228}+0.1400 & \cellcolor[RGB]{251,237,239}+0.0873 & \cellcolor[RGB]{248,221,223}+0.1694 & \cellcolor[RGB]{251,237,238}+0.0895 \\
    31 & \cellcolor[RGB]{221,230,238}-0.3368 & \cellcolor[RGB]{208,221,232}-0.4654 & \cellcolor[RGB]{254,254,254}+0.0018 & \cellcolor[RGB]{224,232,240}-0.3027 & \cellcolor[RGB]{248,221,223}+0.1696 & \cellcolor[RGB]{234,240,245}-0.2022 \\
    \bottomrule
    \end{tabular}
    }
    \caption{Detailed performance impact of \textit{leave-one-out} pruning. Background color intensity indicates the magnitude of impact (\textcolor[RGB]{218,59,70}{red} for increase, \textcolor[RGB]{154,183,205}{blue} for decrease).}
    \label{tab:loo_heatmap}
\end{table}

\section{Quantitative Analysis of Pilot Study}
\label{app:heatmap_data}

In Figure~\ref{fig:heatmap}, we visualized the impact of pruning individual LoRA modules via a heatmap. In this section, we provide the specific numerical values corresponding to that analysis. Table~\ref{tab:loo_heatmap} lists the change in performance ($\Delta$ Acc) when the specific module (layer $l$, task $t$) is removed from the aggregation. 

\section{Implementation Details}
\label{app:implementation}

\subsection{Baseline Formulations}
\label{app:baseline_formulations}

In this section, we formulate the baseline methods using a unified notation. Let $\{\Delta \mathbf{W}_t\}_{t=1}^T$ denote the set of task-specific LoRA updates. The merged update $\Delta \mathbf{W}_{\text{merged}}$ is computed as follows:

\paragraph{Task Arithmetic (TA)}
TA~\cite{ilharco2023editing} assumes that task vectors are independent and constructive. It computes the merged update as a simple weighted sum:
\begin{equation}
    \Delta \mathbf{W}_{\text{TA}} = \lambda \sum_{t=1}^{T} \Delta \mathbf{W}_t,
\end{equation}
where $\lambda$ is a scalar hyperparameter scaling the aggregate strength.

\paragraph{TIES-Merging}
TIES~\cite{yadav2023ties} mitigates interference via a ``Trim, Elect, and Merge'' pipeline. First, it \textit{trims} the task updates to retain only the top-$k$\% largest magnitude parameters, yielding sparse updates $\Delta \hat{\mathbf{W}}_t$. Second, it resolves conflicts by \textit{electing} a unified sign vector $\mathbf{s}^*$ based on the total magnitude of parameters. Finally, it computes a \textit{disjoint mean} by averaging only the values that align with the elected sign:
\begin{equation}
    \small
    \Delta \mathbf{W}_{\text{TIES}} = \lambda \cdot \text{Mean}\left( \{ \Delta \hat{\mathbf{W}}_t \mid \text{sgn}(\Delta \hat{\mathbf{W}}_t) = \mathbf{s}^* \}_{t=1}^T \right),
\end{equation}
where the mean ignores zero entries pruned in the trimming step.

\paragraph{DARE}
DARE~\cite{yu2024language} exploits the extreme redundancy in delta parameters. Operating on the premise that most fine-tuned updates can be removed without performance loss, it employs a stochastic ``Drop And REscale'' strategy. For each task $t$, DARE generates a binary mask $\mathbf{M}_t \sim \text{Bernoulli}(1-p)$ to randomly drop a fraction $p$ of the elements, and rescales the surviving parameters to preserve the expected value of the updates:
\begin{equation}
    \Delta \tilde{\mathbf{W}}_t = \frac{1}{1-p} (\Delta \mathbf{W}_t \odot \mathbf{M}_t).
\end{equation}
These sparsified updates are then typically aggregated via summation or combined with other techniques (e.g., TIES).

\begin{table*}[htbp!]
    \centering
    \small
    \renewcommand{\arraystretch}{1.1} 
    \setlength{\tabcolsep}{5pt}
    \begin{tabular}{l ccc c ccc}
    \toprule
    & \multicolumn{3}{c}{\textbf{NLP (Llama-3)}} & & \multicolumn{3}{c}{\textbf{Vision (ViT-B/32)}} \\
    \cmidrule(lr){2-4} \cmidrule(lr){6-8}
    \textbf{Method} & \textbf{$\lambda$} & \textbf{Top-$k$ (\%)} & \textbf{$p$} & & \textbf{$\lambda$} & \textbf{Top-$k$ (\%)} & \textbf{$p$} \\
    \midrule
    Task Arithmetic (TA) & 0.3 & - & - & & 0.1 & - & - \\
    TIES-Merging & 1.2 & 80 & - & & 0.3 & 40 & - \\
    DARE-TIES & 1.1 & 80 & 0.1 & & 0.3 & 20 & 0.1 \\
    TSV & 0.6 & - & - & & 0.3 & - & - \\
    KnOTS (w/ TIES) & 1.1 & 90 & - & & 0.6 & 90 & - \\
    CoreSpace & 0.5 & - & - & & 0.9 & - & - \\
    \bottomrule
    \end{tabular}
    \caption{Detailed hyperparameter configurations for baselines across NLP and CV benchmarks.
    $\lambda$ denotes the scaling factor (consistent with Eq.~\ref{eq:standard_merge}), Top-$k$ indicates the percentage of parameters retained (density), and $p$ represents the drop rate for DARE.}
    \label{tab:detailed_hyperparams}
\end{table*}

\paragraph{Task Singular Vectors (TSV)}
TSV~\cite{gargiulo2025task} approaches model merging by analyzing the geometric structure of layer-wise weight updates. Recognizing that task matrices are inherently low-rank, TSV decomposes each update via Singular Value Decomposition (SVD) and retains only the most significant singular components, $\Delta \mathbf{W}_t \approx \mathbf{U}_t \mathbf{\Sigma}_t \mathbf{V}_t^\top$. To resolve task interference, it concatenates the singular vectors across all tasks and applies a whitening transformation (formulated as an orthogonal Procrustes problem) to decorrelate them. The merged update is reconstructed from these orthogonalized bases without requiring additional scaling coefficients:
\begin{equation}
    \Delta \mathbf{W}_{\text{TSV}} = \mathbf{U}_{\perp} \mathbf{\Sigma}_{\text{block}} \mathbf{V}_{\perp}^\top,
\end{equation}
where $\mathbf{\Sigma}_{\text{block}} = \text{diag}(\mathbf{\Sigma}_1, \dots, \mathbf{\Sigma}_T)$, and $\mathbf{U}_{\perp}$ (similarly for $\mathbf{V}_{\perp}$) is the orthogonal matrix that minimizes the projection error $\|\mathbf{U}_{\text{cat}} - \mathbf{U}_{\perp}\|_F$ relative to the concatenated singular vectors $\mathbf{U}_{\text{cat}}$.

\begin{table*}[t!]
    \centering
    \resizebox{\linewidth}{!}{
    \begin{tabular}{l llllll l c}
    \toprule
    \textbf{Method} & \textbf{SNLI} & \textbf{MNLI} & \textbf{SICK} & \textbf{QNLI} & \textbf{RTE} & \textbf{SciTail} & \textbf{Avg.} & \textbf{$\Delta$} \\
    \midrule
    Individual Task & 92.50 & 90.31 & 91.58 & 94.49 & 89.86 & 96.52 & 92.54 & - \\
    \midrule
    TA & 86.57 & \textbf{86.06} & 80.60 & 64.94 & 89.13 & 93.32 & 83.44 & - \\
    \textbf{TA + ENMP} & \textbf{88.74} $\pm$ 0.26 & 85.18 $\pm$ 0.39 & \textbf{83.88} $\pm$ 1.63 & \textbf{75.97} $\pm$ 0.57 & \textbf{91.06} $\pm$ 0.42 & \textbf{93.98} $\pm$ 0.12 & \textbf{86.47} $\pm$ 0.25 & \textbf{+3.03} \\
    \midrule
    TIES & 87.74 & 87.34 & 73.99 & 67.60 & \textbf{89.86} & 92.66 & 83.20 & - \\
    \textbf{TIES + ENMP} & \textbf{89.33} $\pm$ 0.25 & \textbf{88.71} $\pm$ 0.56 & \textbf{84.92} $\pm$ 0.75 & \textbf{89.72} $\pm$ 0.63 & 89.37 $\pm$ 0.42 & \textbf{93.04} $\pm$ 0.40 & \textbf{89.18} $\pm$ 0.42 & \textbf{+5.98} \\
    \midrule
    DARE & 87.41 & 87.48 & 71.14 & 68.44 & 87.68 & 92.80 & 82.49 & - \\
    \textbf{DARE + ENMP} & \textbf{88.88} $\pm$ 0.54 & \textbf{88.30} $\pm$ 0.64 & \textbf{84.56} $\pm$ 0.61 & \textbf{89.71} $\pm$ 0.70 & \textbf{88.65} $\pm$ 1.11 & \textbf{93.84} $\pm$ 0.25 & \textbf{88.99} $\pm$ 0.12 & \textbf{+6.50} \\
    \midrule
    TSV & 88.21 & \textbf{85.91} & 81.37 & 72.59 & \textbf{91.30} & 94.17 & 85.59 & - \\
    \textbf{TSV + ENMP} & \textbf{89.22} $\pm$ 0.69 & 85.33 $\pm$ 0.63 & \textbf{86.45} $\pm$ 2.03 & \textbf{87.11} $\pm$ 0.82 & 89.86 $\pm$ 1.25 & \textbf{94.18} $\pm$ 0.26 & \textbf{88.69} $\pm$ 0.33 & \textbf{+3.10} \\
    \midrule
    KnOTS & 82.59 & 84.96 & 82.12 & 79.02 & 90.58 & \textbf{93.98} & 85.54 & - \\
    \textbf{KnOTS + ENMP} & \textbf{87.98} $\pm$ 0.35 & \textbf{88.03} $\pm$ 1.10 & \textbf{88.47} $\pm$ 0.81 & \textbf{90.68} $\pm$ 1.28 & \textbf{91.06} $\pm$ 1.10 & 93.85 $\pm$ 0.43 & \textbf{90.01} $\pm$ 0.27 & \textbf{+4.47} \\
    \midrule
    CoreSpace & 88.65 & \textbf{86.46} & 81.74 & 79.34 & \textbf{92.03} & 94.45 & 87.11 & - \\
    \textbf{CoreSpace + ENMP} & \textbf{90.24} $\pm$ 0.02 & 84.32 $\pm$ 0.71 & \textbf{87.92} $\pm$ 0.64 & \textbf{88.53} $\pm$ 1.19 & 91.06 $\pm$ 0.42 & \textbf{94.97} $\pm$ 0.45 & \textbf{89.51} $\pm$ 0.13 & \textbf{+2.40} \\
    \bottomrule
    \end{tabular}
    }
    \caption{Absolute Accuracy (\%) on NLP Benchmarks. We report the mean accuracy $\pm$ standard deviation across 3 seeds. $\Delta$ denotes the average improvement over the corresponding baseline. The \textit{Individual Task} row represents the performance of models fine-tuned on single tasks.}
    \label{tab:abs_nlp}
\end{table*}
\begin{table*}[ht!]
    \centering
    \resizebox{\linewidth}{!}{
    \begin{tabular}{l llllllll l c}
    \toprule
    \textbf{Method} & \textbf{Cars} & \textbf{DTD} & \textbf{EuroSAT} & \textbf{GTSRB} & \textbf{MNIST} & \textbf{RESISC45} & \textbf{SUN397} & \textbf{SVHN} & \textbf{Avg.} & \textbf{$\Delta$} \\
    \midrule
    Individual Task & 74.00 & 58.30 & 99.00 & 92.70 & 99.30 & 88.40 & 64.50 & 96.20 & 84.05 & - \\
    \midrule
    TA & \textbf{60.67} & 42.98 & 48.81 & \textbf{39.09} & 52.64 & 63.19 & 61.53 & 39.65 & 51.07 & - \\
    \textbf{TA + ENMP} & 60.39 $\pm$ 0.48 & \textbf{43.83} $\pm$ 0.44 & \textbf{54.47} $\pm$ 0.46 & 38.18 $\pm$ 1.38 & \textbf{57.68} $\pm$ 1.95 & \textbf{63.51} $\pm$ 0.19 & \textbf{61.98} $\pm$ 0.13 & \textbf{41.02} $\pm$ 0.92 & \textbf{52.63} $\pm$ 0.31 & \textbf{+1.56} \\
    \midrule
    TIES & \textbf{61.12} & 42.66 & 50.26 & 34.01 & 57.08 & 61.59 & 61.50 & 41.34 & 51.19 & - \\
    \textbf{TIES + ENMP} & 60.33 $\pm$ 0.30 & \textbf{43.28} $\pm$ 0.54 & \textbf{57.58} $\pm$ 2.20 & \textbf{35.16} $\pm$ 0.49 & \textbf{65.12} $\pm$ 0.72 & \textbf{62.31} $\pm$ 0.20 & \textbf{61.88} $\pm$ 0.34 & \textbf{45.04} $\pm$ 2.31 & \textbf{53.84} $\pm$ 0.26 & \textbf{+2.65} \\
    \midrule
    DARE & \textbf{61.00} & 42.55 & 49.41 & 34.90 & 56.31 & 61.65 & 61.34 & 42.81 & 51.25 & - \\
    \textbf{DARE + ENMP} & 60.71 $\pm$ 0.16 & \textbf{44.04} $\pm$ 0.32 & \textbf{58.30} $\pm$ 1.99 & \textbf{36.42} $\pm$ 0.39 & \textbf{66.43} $\pm$ 0.31 & \textbf{62.69} $\pm$ 0.42 & \textbf{61.99} $\pm$ 0.08 & \textbf{45.78} $\pm$ 1.44 & \textbf{54.54} $\pm$ 0.04 & \textbf{+3.29} \\
    \midrule
    TSV & \textbf{61.85} & 44.00 & 52.07 & 41.58 & 59.11 & 64.87 & 61.44 & 47.42 & 54.04 & - \\
    \textbf{TSV + ENMP} & 61.23 $\pm$ 0.34 & \textbf{44.01} $\pm$ 1.06 & \textbf{61.94} $\pm$ 3.77 & \textbf{43.07} $\pm$ 0.79 & \textbf{69.01} $\pm$ 1.74 & \textbf{65.33} $\pm$ 0.80 & \textbf{62.06} $\pm$ 0.27 & \textbf{51.10} $\pm$ 0.53 & \textbf{57.22} $\pm$ 0.20 & \textbf{+3.18} \\
    \midrule
    KnOTS & 61.23 & 42.34 & 46.59 & 40.91 & 61.16 & 62.98 & 60.36 & 47.35 & 52.87 & - \\
    \textbf{KnOTS + ENMP} & \textbf{61.40} $\pm$ 0.25 & \textbf{44.10} $\pm$ 0.09 & \textbf{54.50} $\pm$ 0.72 & \textbf{41.49} $\pm$ 1.66 & \textbf{79.90} $\pm$ 1.52 & \textbf{63.09} $\pm$ 1.43 & \textbf{61.54} $\pm$ 0.48 & \textbf{58.60} $\pm$ 2.71 & \textbf{58.08} $\pm$ 0.33 & \textbf{+5.21} \\
    \midrule
    CoreSpace & 61.34 & 49.57 & 52.63 & \textbf{78.15} & 70.50 & 74.56 & 62.90 & 51.51 & 62.64 & - \\
    \textbf{CoreSpace + ENMP} & \textbf{62.26} $\pm$ 0.09 & 48.12 $\pm$ 1.62 & \textbf{66.48} $\pm$ 1.56 & 75.45 $\pm$ 1.44 & \textbf{75.75} $\pm$ 1.24 & \textbf{75.13} $\pm$ 0.56 & \textbf{63.09} $\pm$ 0.30 & \textbf{53.93} $\pm$ 2.03 & \textbf{65.03} $\pm$ 0.11 & \textbf{+2.39} \\
    \bottomrule
    \end{tabular}
    }
    \caption{Absolute Accuracy (\%) on Vision Benchmarks. The improvement ($\Delta$) indicates the gain of ENMP over the baseline method. The \textit{Individual Task} row represents the performance of models fine-tuned on single tasks.}
    \label{tab:abs_cv}
\end{table*}

\paragraph{KnOTS}
KnOTS~\cite{stoica2025model} addresses the misalignment issue in LoRA-finetuned models by projecting task-specific updates into a shared geometric subspace. 
By attributing the poor mergeability of LoRA models to their updates residing in disparate subspaces, KnOTS concatenates the weight updates from all tasks layer-wise and applies Singular Value Decomposition (SVD) to extract a common orthonormal basis $\mathbf{U}$ and a scaling matrix $\mathbf{\Sigma}$. 
This decomposition isolates task-specific variations into the right singular vectors $\mathbf{V}_t$, which are aligned to the shared basis. 
Standard merging algorithms (such as TIES or Task Arithmetic) are then applied exclusively to these aligned vectors to produce a unified component $\mathbf{V}_{\text{merged}}$. 
The final merged update is reconstructed by projecting back via the shared basis:
\begin{equation}
    \Delta \mathbf{W}_{\text{merged}} = \mathbf{U} \mathbf{\Sigma} \mathbf{V}_{\text{merged}}^\top,
\end{equation}
where $\mathbf{V}_{\text{merged}} = \text{Merge}(\{\mathbf{V}_1, \dots, \mathbf{V}_T\})$. 
This procedure effectively aligns the representation spaces of disjoint LoRA models without requiring additional data or gradient-based optimization. 
For the main experiments, we follow the best practices from the original paper, utilizing TIES for the merging stage.

\paragraph{CoreSpace}
CoreSpace~\cite{panariello2025accurate} proposes a computationally efficient framework for merging LoRA-adapted models by operating within a compact, shared geometric subspace. 
Unlike methods that merge in the high-dimensional parameter space or require costly SVD on full weight updates (e.g., KnOTS), CoreSpace constructs a common alignment basis $(\mathbf{U}_B^{\text{ref}}, \mathbf{V}_A^{\text{ref}})$ by decomposing the concatenated low-rank factors $\mathbf{B}_t$ and $\mathbf{A}_t$ across all tasks. 
Each task's update is projected into this space to obtain a dense \textit{Core Matrix} $\tilde{\mathbf{M}}_t$, capturing the task-specific transformation without information loss. 
Merging is performed on these low-dimensional matrices, and the final update is reconstructed via the reference bases:
\begin{equation}
    \Delta \mathbf{W}_{\text{Core}} = \mathbf{U}_B^{\text{ref}} \text{Merge}\left( \{ \tilde{\mathbf{M}}_t \}_{t=1}^T \right) (\mathbf{V}_A^{\text{ref}})^\top,
\end{equation}
where $\tilde{\mathbf{M}}_t = (\mathbf{U}_B^{\text{ref}})^\top \mathbf{B}_t \mathbf{A}_t \mathbf{V}_A^{\text{ref}}$. 
This approach decouples merging complexity from the model dimension, ensuring scalability while theoretically guaranteeing zero reconstruction error relative to full-space concatenation.
For the main experiments, we follow the best practices from the original paper in merging stage, employing TSV for NLP and TSV + Iso-C for the Vision benchmark.

\subsection{Hyperparameter Configurations}
\label{app:hyperparams}

To ensure fair comparison and reproducibility, we align our baseline settings with prior works. We observe that optimal hyperparameters vary significantly between modalities due to differences in backbone architectures (Llama-3 vs. ViT) and task characteristics. Table~\ref{tab:detailed_hyperparams} details the specific configurations for both NLP and Vision benchmarks.

\section{Additional Quantitative Results}
\label{app:additional_results}

In Table~\ref{tab:main_results_nlp} and Table~\ref{tab:main_results_cv}, we reported the Normalized Accuracy to provide a balanced view across tasks with varying difficulty. To ensure transparency and facilitate comparison with future works, we present the \textbf{Absolute Accuracy} for all benchmarks in this section.

\subsection{NLP Benchmark Results}
\label{app:nlp_results}
Table~\ref{tab:abs_nlp} reports the raw accuracy scores for the NLP tasks. 
As demonstrated in Table~\ref{tab:abs_nlp}, our method (ENMP) consistently improves performance across all baselines. It is particularly encouraging to see that even for strong baselines such as TSV, KnOTS, and CoreSpace, which already achieve high base accuracy, ENMP still provides significant further improvements of +3.10\%, +4.47\%, and +2.40\%, respectively.

\subsection{Vision Benchmark Results}
\label{app:cv_results}
Table~\ref{tab:abs_cv} presents the absolute accuracy for the CV tasks using the ViT-B/32 backbone.
Similar to the NLP results, ENMP provides a universal boost across all CV baselines. 
It is particularly effective when combined with subspace-based methods like KnOTS, achieving a remarkable +5.21\% improvement. Even for the state-of-the-art method CoreSpace, which already operates in a highly optimized subspace, ENMP still squeezes out an additional +2.39\% accuracy, demonstrating its complementary nature to existing subspace-based merging techniques.

\end{document}